\documentclass[11pt]{article}

\usepackage{asgard_theme}

\usepackage[T1]{fontenc}
\usepackage[utf8]{inputenc}
\usepackage{lmodern}
\usepackage{microtype}
\usepackage{url}
\usepackage[export]{adjustbox}
\usepackage{xspace}

\usepackage{fvextra}

\usepackage[margin=1in]{geometry}

\usepackage{graphicx}
\usepackage{subcaption}
\usepackage{caption}
\usepackage{longtable}

\usepackage{booktabs}
\usepackage{tabularx}
\usepackage{multirow}
\usepackage{array}

\usepackage{amsmath, amssymb}

\usepackage{enumitem}
\usepackage{float}

\setlist{nosep}

\usepackage{xcolor}
\usepackage[hidelinks]{hyperref}
\usepackage[nameinlink,capitalize]{cleveref}
\usepackage{authblk}
\usepackage{etoolbox}

\makeatletter
\patchcmd{\@maketitle}{\null\vskip 2em}{\vskip -0.5em}{}{}
\makeatother

\usepackage{titlesec}
\titlespacing*{\section}{0pt}{0.9em}{0.4em}
\titlespacing*{\subsection}{0pt}{0.7em}{0.3em}
\titlespacing*{\subsubsection}{0pt}{0.6em}{0.2em}

\newcommand{\benchmark}{AsgardBench\xspace}

\newif\ifdraftdyslexic
\draftdyslexicfalse

\ifdraftdyslexic
  \usepackage{fontspec}
\fi

\title{\benchmark --- Evaluating Visually Grounded Interactive Planning Under Minimal Feedback\protect\footnote{\url{https://github.com/microsoft/AsgardBench}}}
\author{Andrea Tupini}
\author{Lars Liden}
\author{Reuben Tan}
\author{Yu Wang}
\author{Jianfeng Gao}
\affil{Microsoft Research}
\date{}

\begin{document}
\maketitle

\vspace{-2em}
\begin{figure}[H]
    \centering
    \footnotesize
    \renewcommand{\arraystretch}{1.2}
    \setlength{\tabcolsep}{8pt}
    \newcolumntype{Y}{>{\raggedright\arraybackslash}X}
    \begin{tabularx}{\textwidth}{|Y|Y|Y|}
    \hline
    \multicolumn{1}{|c|}{\normalsize\textbf{Initial Observation}} & \multicolumn{1}{c|}{\normalsize\textbf{Observe Cup Dirty}} & \multicolumn{1}{c|}{\normalsize\textbf{Observe Items In Sink}} \\

    {\centering\includegraphics[width=0.75\linewidth]{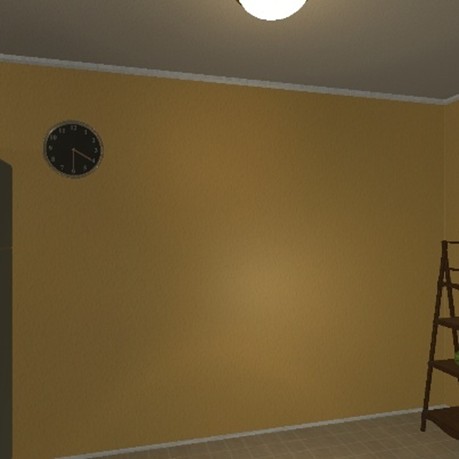}\par} &
    {\centering\includegraphics[width=0.75\linewidth]{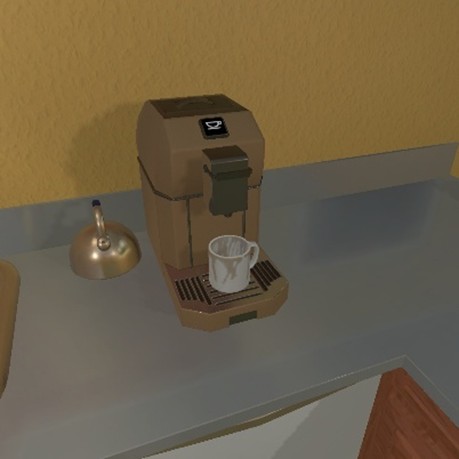}\par} &
    {\centering\includegraphics[width=0.75\linewidth]{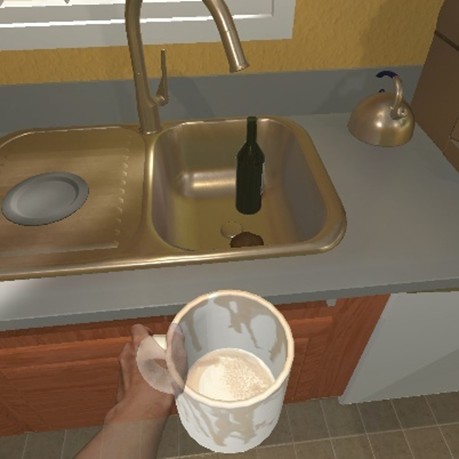}\par} \\

    \vspace{-8pt}
    {\scriptsize
    \begin{enumerate}[nosep, leftmargin=*, align=left]
        \item FIND Mug
        \item PICKUP Mug
        \item FIND CoffeeMachine
        \item PUT CoffeeMachine
        \item TOGGLE\_ON CoffeeMachine
        \item TOGGLE\_OFF CoffeeMachine
        \item PICKUP Mug
        \item DRINK Mug
        \item FIND SinkBasin
        \item ...
    \end{enumerate}
    }
    \vspace{-6pt} &

    \vspace{-8pt}
    {\scriptsize
    \begin{enumerate}[nosep, leftmargin=*, align=left]
        \item FIND Mug (DONE)
        \item PICKUP Mug
        \item FIND SinkBasin
        \item PUT SinkBasin
        \item FIND DishSponge
        \item PICKUP DishSponge
        \item FIND SinkBasin
        \item TOGGLE ON Faucet
        \item CLEAN Mug
        \item ...
    \end{enumerate}
    }
    \vspace{-6pt} &

    \vspace{-8pt}
    {\scriptsize
    \begin{enumerate}[nosep, leftmargin=*, align=left]
        \item FIND Mug (DONE)
        \item PICKUP Mug (DONE)
        \item FIND SinkBasin (DONE)
        \item PUT CounterTop
        \item PICKUP WineBottle
        \item PUT CounterTop
        \item FIND Potato
        \item PICKUP Potato
        \item FIND CounterTop
        \item ...
    \end{enumerate}
    }
    \vspace{-6pt} \\
    \hline

    \end{tabularx}
    \caption{\textbf{Agent observations and corresponding action plans in \benchmark{}.}
    Below each image is the action plan generated from that observation. This illustrates how \benchmark{} requires agents to update or change their plans based on new visual evidence rather than following a fixed
    action sequence.}
    \label{fig:2-assumptions-change-as-agent-discovers-environment}
\end{figure}

\begin{abstract}
With \benchmark we aim to evaluate visually grounded, high-level action sequence generation and interactive planning, focusing specifically on plan adaptation during execution based on visual observations rather than navigation or low-level manipulation. In the landscape of embodied AI benchmarks, \benchmark targets the capability category of interactive planning, which is more sophisticated than offline high-level planning as it requires agents to revise plans in response to environmental feedback, yet remains distinct from low-level execution.

Unlike prior embodied AI benchmarks that conflate reasoning with navigation or provide rich corrective feedback that substitutes for perception, \benchmark restricts agent input to images, action history, and lightweight success/failure signals, isolating interactive planning in a controlled simulator without low-level control noise.

The benchmark contains 108 task instances spanning 12 task types, each systematically varied through object state, placement, and scene configuration. These controlled variations create conditional branches in which a single instruction can require different action sequences depending on what the agent observes, emphasizing conditional branching and plan repair during execution.

Our evaluations of leading vision language models show that performance drops sharply without visual input, revealing weaknesses in visual grounding and state tracking that ultimately undermine interactive planning. Our benchmark zeroes in on a narrower question: can a model actually use what it sees to adapt a plan when things do not go as expected?
\end{abstract}

\section{Introduction}
\label{sec:introduction}

Vision language models have enabled agents to generate plausible action sequences for embodied tasks, prompting renewed interest in benchmarks that evaluate multimodal reasoning in interactive environments. Many existing benchmarks mix high-level reasoning with navigation and other environment-specific mechanics and agents may succeed by exploiting textual priors, memorized procedures, or detailed corrective feedback that substitutes for perception, even when visual grounding is weak. We found that producing a plan upfront is rarely enough for embodied tasks, as agents must continually reassess what they see and adjust accordingly. They must repeatedly interpret visual observations, detect when assumptions no longer hold, and revise action sequences accordingly. Despite its importance, visually grounded interactive planning is rarely evaluated on its own. In several widely used benchmarks, most notably in ALFWorld \cite{ALFWorld20} and EmbodiedBench \cite{yang2025embodiedbench}, text only agents perform competitively or even better, and detailed failure messages often reveal why an action failed, reducing the need for genuine multimodal reasoning.

We designed \benchmark to isolate this gap, deliberately abstracting away navigation and low-level manipulation to avoid conflating reasoning failures with control errors. Recent embodied AI evaluation surveys \cite{embodiedaisurvey, authoreaSurvey} categorize benchmarks along two axes: the capability being measured (e.g., navigation, manipulation, planning) and the evaluation setting (e.g., simulators, datasets, real world environments). Within this taxonomy, \benchmark targets interactive planning as the capability being evaluated and uses a controlled simulator as the evaluation setting. Agents interact with the environment through a simplified action space and receive only lightweight success or failure signals, requiring them to infer task relevant state, such as object cleanliness, container accessibility, or scene configuration, directly from images.  Tasks in \benchmark are constructed so that a single instruction can require different action sequences depending on what the agent observes during execution. This setup requires conditional branching and plan repair based on visual observations, exposing whether agents can maintain coherent state across steps and adapt plans when observations contradict expectations. Through systematic ablations, we show that visual grounding is essential for success in \benchmark, clarifying where current multimodal agents struggle and distinguishing visually grounded plan adaptation from broader embodied AI benchmarks where multimodal perception can often be bypassed.

\subsection{Contributions}
\label{subsec:contributions}

Our work makes the following contributions:

\begin{itemize}

    \item \textbf{A benchmark that isolates visually grounded interactive planning.} It removes navigation, low-level manipulation, and rich corrective feedback so the evaluation centers on whether agents can ground action sequencing in visual observations. Because tasks require multiple steps with conditional dependencies, agents must not only react to what they see but also maintain coherent progress toward their goal---tracking intent across steps rather than treating each observation in isolation.

    \item \textbf{Controlled task variations that require conditional branching during execution.} Each instruction admits multiple valid action sequences depending on object cleanliness, accessibility, or scene configuration, enabling precise evaluation of visual grounding, state tracking, and adaptive plan revision.

    \item \textbf{A targeted empirical analysis of state-of-the-art multimodal models within this capability category.} Through image based, text only, and feedback ablated conditions, we reveal consistent failures in visual discrimination, internal state maintenance, and interactive planning, highlighting gaps that are not exposed by broader embodied AI benchmarks.
\end{itemize}

\section{Related Work}
\label{sec:related-work}

Embodied AI benchmarks have traditionally evaluated agents on end-to-end task execution that combines navigation, manipulation, and language grounding. Early benchmarks like ALFRED \cite{ALFRED20} and its photo-realistic variant, ReALFRED \cite{kim2024realfred}, frame household activities as long horizon instruction following problems in simulated 3D environments. These benchmarks have advanced multimodal agents, but their integrated design makes it difficult to isolate high-level reasoning failures from errors in navigation, viewpoint selection, or low-level interaction. Empirical results, particularly in ALFWorld \cite{ALFWorld20} and as reflected in EmbodiedBench \cite{yang2025embodiedbench}, show that agents can often perform competitively using textual priors or memorized action templates, even when visual grounding is weak or absent.

Large scale activity benchmarks, for example BEHAVIOR-1K \cite{li2024behavior1k}, extend this paradigm by increasing task diversity and physical realism, emphasizing general manipulation competence across 1,000 everyday activities. While these environments stress robustness and scalability, performance is often strongly influenced by challenges in low-level control and physics, which can obscure deficiencies in interactive decision making and plan adaptation. Similarly, procedurally generated extensions of AI2-THOR \cite{ai2thor} like ProcTHOR \cite{procthor} increase visual diversity but retain a focus on navigation and interaction mechanics.

Meta-benchmarks (i.e., EmbodiedBench \cite{yang2025embodiedbench}) attempt to broaden evaluation coverage by aggregating tasks across multiple environments and organize them by capability categories. This approach improves comparability across models but inherits many of the design assumptions of their source benchmarks, including varying degrees of navigation complexity and simulator specific affordances. As a result, while these suites provide comprehensive evaluation across multiple dimensions, the varying feedback mechanisms across component benchmarks can make it challenging to isolate interactive planning as a distinct capability.

Other work examines planning in offline settings, where agents generate full action sequences without interacting with the environment. ET-PLAN-BENCH \cite{zhang2024et-plan-bench} evaluates whether agents can generate valid plans with spatiotemporal reasoning, though they often provide structured task representations or environmental feedback that may reduce the demands on continuous real time visual state estimation during execution. LoTa-Bench \cite{choi2024lotaBench} and related frameworks examine long horizon planning with language oriented task decomposition, though the emphasis on symbolic representations may reduce the need for visual grounding. ActPlan-1K \cite{su2024actplan1kbenchmark} emphasize procedural plan generation from video observations but evaluate plans in offline settings without requiring agents to revise them during execution.

A related line of work evaluates agents in dynamic, stepwise environments. WebArena \cite{webarena2023}, WebShop \cite{webshop2022} and OSWorld \cite{osworld2023}. These settings share with \benchmark the requirement that an agent operate interactively in a simulator, making them conceptually aligned with the broader goal of evaluating interactive planning. However, web based and desktop control environments differ in several important ways. In particular their action spaces involve symbolic clicking, typing, or selecting UI elements. \benchmark employs a fully controlled 3D simulator with unstructured RGB observations.

Other diagnostic frameworks, including CanDo! \cite{chia2024candodatasetneurosymbolicgrounded} and the Embodied Agent Interface (EAI) \cite{li2024embodied-agent-interface}, aim to decompose embodied decision making into modular components and provide fine grained error analysis. While valuable for system level diagnosis, these approaches often rely on structured state, explicit affordances, or detailed error signals, which can mask the extent to which agents genuinely rely on visual input during execution.

Unlike settings where text only agents succeed by exploiting structured feedback, \benchmark requires models to interpret the scene itself, as confirmed by systematic ablations that remove or degrade visual input. By varying object states and placements, providing only minimal feedback, and requiring interactive execution, \benchmark exposes limitations in models' ability to maintain coherent state, detect mismatches between expectation and observation, and adapt plans mid-execution. This positions \benchmark as the first benchmark to isolate interactive planning in embodied environments, filling a gap between physically rich embodied simulations and abstract offline planning evaluations.
\section{Benchmark Design}
\label{sec:benchmark-design}

In order to focus on the ability of models to adapt plans as new visual information becomes available during execution, skills such as navigation and grasping are deliberately excluded, as benchmarks that emphasize navigation or manipulation can make it difficult to disentangle high-level reasoning failures from low-level control errors, as seen in embodied evaluations derived from ALFRED \cite{ALFRED20}, BEHAVIOR-1K \cite{li2024behavior1k}, and AI2-THOR-based tasks \cite{ai2thor}. Consistent with the capability-focused evaluation principles discussed in \autoref{sec:introduction} \cite{embodiedaisurvey, authoreaSurvey}, \benchmark adopts a controlled simulation based design that targets interactive planning while minimizing confounding factors from navigation and low-level control.

\benchmark was built on AI2-THOR, an open source, mature simulation environment widely used in embodied AI research \cite{ai2thor}. \benchmark uses three of AI2-THOR's scene types: \textbf{Kitchens} (for cooking and organization tasks), \textbf{Living Rooms} (for simple interaction tasks), and \textbf{Bathrooms} (for cleaning tasks). For the purposes of the \benchmark benchmark, a higher-level action abstraction layer was added, simplifying several elements for the agent so that the focus remains on visually grounded interactive planning and not motor control.

In AI2-THOR, to interact with an object (e.g., PUT or OPEN), an agent must navigate to the object and identify a position relative to it from which the action can be executed. For example, when opening a fridge, the agent can be neither too close nor too far from the door. \benchmark simplifies this by automatically placing the agent in a position from which it can interact with the object, including setting whether the agent is crouched or upright. In this way, we remove navigation entirely by placing the agent directly in front of the relevant object, ensuring that errors reflect planning difficulties. For all object interactions, however, we still require that the object be within the agent's field of view before any interaction can occur. A \texttt{"FIND {object}"} action was also added, which positions the object within the agent's field of view, eliminating the need for visual search and navigation while still requiring the agent to interpret what it sees and update its plan accordingly.

The sizes of containers and objects in the AI2-THOR environment vary, meaning that not every container will fit an object of a particular type. For example, a Bowl may fit in only a subset of the available Cabinets. This has been abstracted in the simulation environment so that the agent does not need to determine which container of a particular type to employ. When the agent requests an "OPEN Drawer" or "PUT Drawer" action, \benchmark automatically selects a drawer that fits the object. Likewise, when the agent performs a "PUT CounterTop" whilst holding an item, \benchmark selects a CounterTop that has sufficient space for the held item, removing low-level affordance reasoning that is not central to the benchmark's goals and keeping the evaluation focused on interactive planning.

\subsection{Tasks}
\label{subsec:tasks}

A Task in \benchmark is defined by a high-level instruction such as "\textit{Make a slice of toast and serve it on a plate}." For each Task instruction, we generated variations by changing the initialization of Scene objects. See \autoref{tab:1-task-variations} for a list of Task instruction types and their associated variations. Because of these variations, the steps the agent must take to complete the Task vary, and a single instruction can branch into multiple execution paths depending on what the agent observes during the episode. \autoref{fig:coffee-plans} in the appendix illustrates how these variations cause a single Task instruction to diverge into different action plans, creating controlled conditional branches.

\begin{table}[H]
\centering
\small
\renewcommand{\arraystretch}{1.5}
\begin{tabularx}{\textwidth}{|>{\raggedright\arraybackslash}p{4.2cm}|c|c|>{\raggedright\arraybackslash}X|}
\hline
\multicolumn{1}{|c|}{\normalsize\textbf{Task Type}} & {\normalsize\textbf{Scene Type}} & {\normalsize\textbf{\# Tasks}} & \multicolumn{1}{c|}{\normalsize\textbf{Task Variations}} \\
\hline

``Consume coffee from a mug, then wash and store the mug'' & Kitchen & 9 &
\vspace{-6pt}
\begin{itemize}[nosep, leftmargin=*]
  \item Vary whether the mug is clean or dirty
  \item Vary the initial fill state of the mug (empty, has coffee)
\end{itemize}
\vspace{-6pt} \\
\hline

``Microwave a potato and serve it in a bowl'' & Kitchen & 6 &
\vspace{-6pt}
\begin{itemize}[nosep, leftmargin=*]
  \item Requires a bowl to serve
  \item Vary whether the bowl is dirty
\end{itemize}
\vspace{-6pt} \\
\hline

``Cook an egg in a pan and serve it on a plate'' & Kitchen & 12 &
\vspace{-6pt}
\begin{itemize}[nosep, leftmargin=*]
  \item Requires a pan to cook and a plate to serve
  \item Vary whether the pan and plate are dirty/clean
\end{itemize}
\vspace{-6pt} \\
\hline

``Make a slice of toast and serve it on a plate'' & Kitchen & 6 &
\vspace{-6pt}
\begin{itemize}[nosep, leftmargin=*]
  \item Vary whether the plate is clean or dirty
\end{itemize}
\vspace{-6pt} \\
\hline

``Fry a potato slice and serve it on a plate'' & Kitchen & 12 &
\vspace{-6pt}
\begin{itemize}[nosep, leftmargin=*]
  \item Vary whether some or all items are clean or dirty
\end{itemize}
\vspace{-6pt} \\
\hline

``Put away the \{Food\}'' & Kitchen & 3 &
\vspace{-6pt}
\begin{itemize}[nosep, leftmargin=*]
  \item Vary which items are clean/dirty
\end{itemize}
\vspace{-6pt} \\
\hline

``Put away the \{Dishes\}'' & Kitchen & 15 & \\
\hline

``Put away the \{Silverware\}'' & Kitchen & 3 & \\
\hline

``Slice the \{Lettuce / Apple / Tomato\} and put a piece in the \{Plate / Bowl / Pan / Pot\}'' & Kitchen & 27 &
\vspace{-6pt}
\begin{itemize}[nosep, leftmargin=*]
  \item Vary which dishes are clean/dirty
\end{itemize}
\vspace{-6pt} \\
\hline

``Clear the dining table and set it with a mug of coffee and a slice of toast'' & Kitchen & 9 &
\vspace{-6pt}
\begin{itemize}[nosep, leftmargin=*]
  \item Vary whether the mug is dirty
  \item Vary whether the mug is empty or contains coffee
\end{itemize}
\vspace{-6pt} \\
\hline

``Clean the mirror'' & Bathroom & 3 & \\
\hline

``Turn on the television'' & Living Room & 3 & \\
\hline

\end{tabularx}
\caption{Twelve Task types are included in the \benchmark{} benchmark. The \textit{\# Tasks} column indicates the number of task instances in the 108-task test set for each type. In addition to the task-specific variations, object locations were randomized for all Task types, including the contents of the SinkBasin for Kitchen scenes.}
\label{tab:1-task-variations}
\end{table}

To create meaningful variations, we designed each Task and its variations so that the required action sequence diverges based on visual observations made during execution, ensuring that interactive plan adaptation is necessary for success. For example, in \autoref{fig:2-assumptions-change-as-agent-discovers-environment} (left panel), the agent forms an initial plan for "\textit{Consume coffee from a mug, then wash and store the mug}" that includes finding the mug and then putting it in the CoffeeMaker. However, upon observing that the Mug is dirty, it must alter the plan to include washing the Mug first (center panel). Subsequently, the agent observes that the SinkBasin contains other objects, and it must further alter its plan to remove items from the SinkBasin before washing the Mug (right panel).

Concretely, at each turn the model produces a complete action sequence covering all remaining steps, but only the first action in that sequence is executed. After execution, the agent receives updated images and minimal success/failure feedback, and the previous plan is carried forward for re-evaluation. This loop ensures that the model must continuously reconcile its plan with what it actually observes, rather than committing to a fixed sequence up front. Full details of the observation space and prompt structure are provided in \autoref{apdx:observation-space}.

By requiring the agent to revise its plan dynamically during execution, the benchmark ensures that the agent must ground its planning in the visual environment. \autoref{fig:6-baseline-vs-detailed-errors} shows that even with detailed feedback on why actions fail, the agent's performance drops dramatically if it does not have access to visual observations. Prior work on derived text only abstractions of embodied benchmarks, most notably ALFWorld \cite{ALFWorld20} for ALFRED \cite{ALFRED20}, shows that agents achieve high success rates in the text only environment but experience severe degradation when transferred to the embodied setting. More broadly, reducing control complexity consistently improves performance across embodied settings: hierarchical skill decompositions substantially outperform flat policies in physics-based manipulation \cite{szot2021habitat}, and oracle stopping conditions boost success in vision-and-language navigation \cite{anderson2018vln}, further underscoring how dependent current agents remain on simplified action abstractions.

\subsection{Evaluation Metrics}
\label{subsec:evaluation-metrics}

A Task is considered to have been completed by the Agent when all task specific goals have been met. For example, the \textit{Coffee has been drunk, the Mug is clean, and the Mug has been placed in a cabinet}. See \autoref{tab:task-specific-goals} for all task specific goals. Additionally, for the kitchen scene, the benchmark requires that the world be left in a reasonable state (i.e., the Refrigerator is not open and the Faucet is not running), mirroring cleanup constraints used in prior embodied benchmarks \cite{ALFRED20, ai2thor}.

These metrics are designed to evaluate visually grounded interactive planning: the agent must not only achieve the final goals but must do so by adapting its plan based on visual observations made during execution. Because different scene variations require different action sequences, success reflects the agent's ability to detect visually observable state, maintain internal task progress, and revise its plan accordingly.

\subsection{Termination Conditions}
\label{subsec:termination-conditions}

Determining when to terminate a test is not straightforward. Agents should be given adequate opportunity to solve a Task, yet time and compute cycles should not be wasted when the Agent is floundering. The easiest Tasks can be completed in only four steps, but in some cases the Agent may take over 100 steps to reach the goal state, a pattern also observed in long-horizon embodied evaluations such as ALFRED and BEHAVIOR-1K \cite{ALFRED20, li2024behavior1k}.

For \benchmark, testing for a Task is terminated under the following conditions, balancing exploration with practical compute limits:

\begin{enumerate}
    \item All goals are met.
    \item The Agent attempts ten undoable actions in a row.
    \item The Agent repeats the same action or sequence of actions more than eight times.
    \item The Agent exceeds the maximum allowed number of steps (soft or hard limits).
\end{enumerate}

On the last point, the benchmark imposes soft and hard maxima on the total number of steps that an Agent is allowed to use. The soft limit is 1.5x the number of steps in the known solution created by the plan generator (see the Benchmark Generation section in \ref{apdx:benchmark-generation}), with a minimum of 15 steps. The hard limit is 2x the number of steps, with a minimum of 20 steps. When the Agent reaches the soft limit, it is allowed to continue as long as it has interacted with an object that was not employed by it in the previous ten steps or has completed a new subgoal, until it reaches the hard limit. This novelty check allows an Agent to continue if it is trying new actions but helps terminate agents that have become stuck.

We chose these termination thresholds empirically by examining Agent trajectories. It remains possible that an Agent might eventually reach its goals given additional steps, but we believe the current approach keeps a reasonable balance between allowing recovery from errors and preventing unbounded exploration.  These conditions also ensure that evaluation remains focused on interactive planning: agents are given enough room to revise their plans in response to new visual information, but not so much freedom that uncontrolled exploration obscures whether plan adaptation is occurring.

\section{Results}
\label{sec:results}

We tested leading frontier and near frontier proprietary and open source models with images and in \textit{Text Only} mode. For cost control, some of the more expensive models were not included in all experimental conditions.   Following the "\textit{what to evaluate / where to evaluate}" framework \cite{embodiedaisurvey, authoreaSurvey}, these results speak directly to the interactive planning capability category.  Success depends on adapting plans based on what the agent sees, not on producing a fixed plan or relying on detailed feedback.

\begin{figure}
    \centering
    \includegraphics[width=0.8\linewidth]{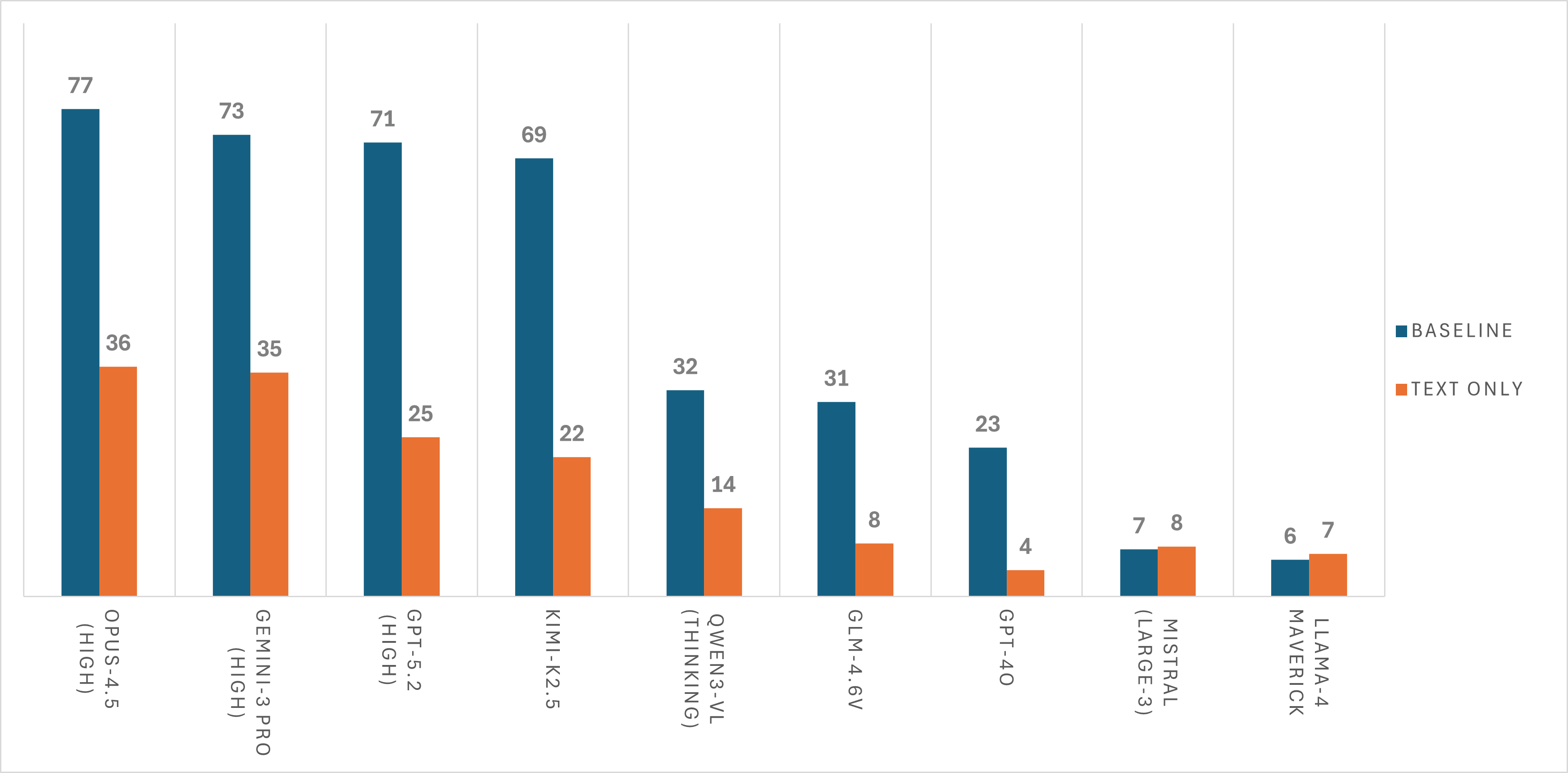}
    \caption{\textbf{Success rates for each model under image based and \textit{Text-Only} conditions.}
Visual input substantially improves performance for all but the weakest models, confirming that \benchmark requires perception-conditioned reasoning.  Agents can't rely on memorized action templates or detailed feedback. \textit{Text-Only} performance remains low across models, in contrast to prior embodied benchmarks where \textit{Text-Only} agents can perform competitively.}
    \label{fig:4-model-performance}
\end{figure}

\begin{figure}
    \centering
    \includegraphics[width=0.8\linewidth]{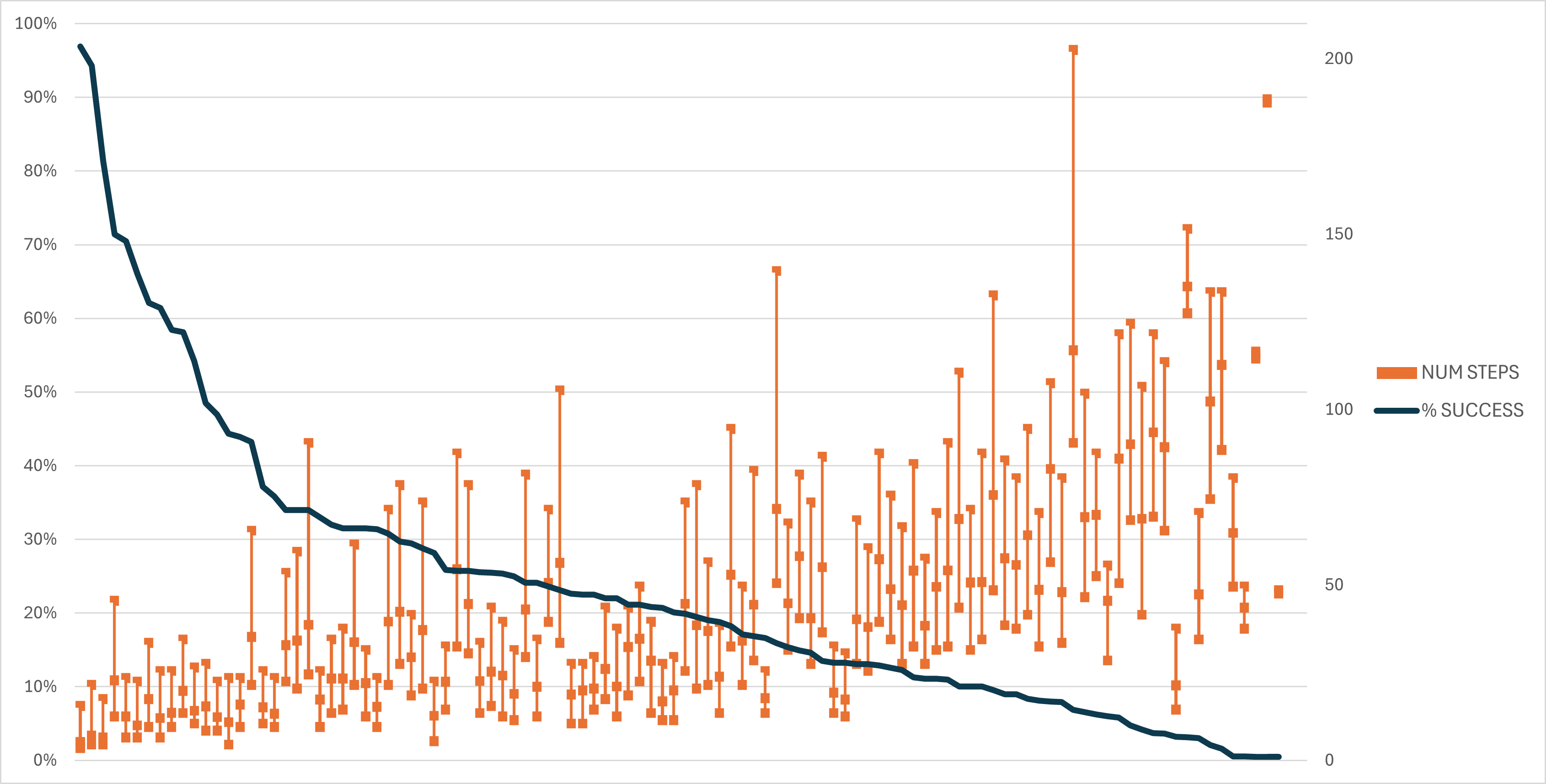}
    \caption{\textbf{Relationship between task difficulty and the range of steps required for successful completion.}
We found that tasks that require longer or more variable action sequences tend to have lower success rates, indicating that models struggle with long-horizon dependencies and conditional branching. Each point represents one of the 108 tasks in \benchmark.}
    \label{fig:5-number-of-steps-vs-success-rate}
\end{figure}

As shown in \autoref{fig:4-model-performance}, all but the lowest performing models benefit from visual input, more than doubling performance over the \textit{Text-Only} condition. This indicates that success on \benchmark requires visual grounding and cannot be achieved through memorized action templates or static plans. This stands in contrast to derived text only abstractions of embodied benchmarks, like ALFWorld \cite{ALFWorld20} for ALFRED \cite{ALFRED20}, where agents can perform competitively by exploiting textual priors or structured feedback.  We also examined the range of steps required to successfully complete each Task. \autoref{fig:5-number-of-steps-vs-success-rate} shows the range of steps plotted against the average success rate for each of the 108 tasks in the benchmark.  Tasks that require longer action sequences tend to be more difficult, a pattern also observed in long horizon embodied evaluations derived from ALFRED \cite{ALFRED20} and in large scale activity benchmarks like BEHAVIOR-1K \cite{li2024behavior1k}.  Longer horizon tasks in \benchmark typically involve more assumption changing variations, increasing the need for interactive plan adaptation.

In the \textit{Baseline} configuration, the prompt included a list of the Agent's previous actions and an indication of whether each attempted action was successful (\textit{Success/Failure}). Two additional variations were examined: removing feedback about the success or failure of each action entirely (\textit{No Feedback}) and providing detailed feedback that explains why an action failed (\textit{Detailed Feedback}). See \autoref{tab:2-comparing-past-action-prompts-different-feedbacks} for a demonstration of how the prompt varies by feedback type.

In the \textit{No Feedback} condition, performance degrades because the agent must rely solely on visual feedback to determine whether an action has succeeded. This reliance is challenging in AI2-THOR \cite{ai2thor}, as visual differences between states can be extremely subtle (for example, the on/off state of a Toaster or Microwave).

\begin{figure}
    \centering
    \includegraphics[width=0.8\linewidth]{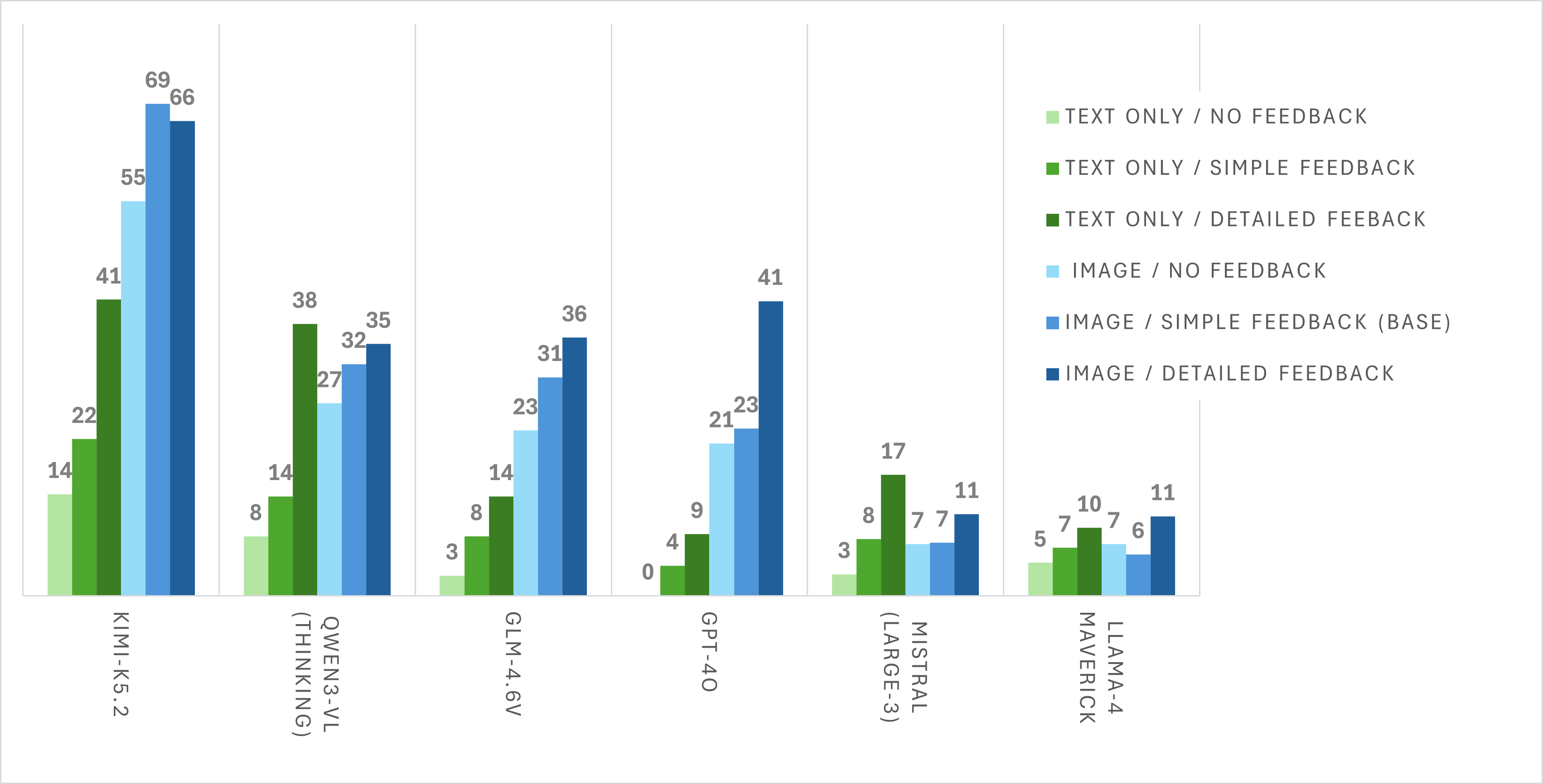}
    \caption{\textbf{Effect of feedback type on model performance.}
Removing success/failure signals (\textit{No Feedback}) reduces accuracy, while providing detailed error messages (\textit{Detailed Feedback}) sharply increases performance, including for \textit{Text-Only} agents.}
    \label{fig:6-baseline-vs-detailed-errors}
\end{figure}

\textit{Detailed Feedback} substantially reduces the need for visual grounding, as the feedback provides explicit corrective information that could, in principle, guide the agent toward the correct action sequence. \textit{Detailed Feedback} does raise performance across all models. For some models---notably Qwen3-VL, Mistral-Large-3, and Maverick---\textit{Text-Only} with \textit{Detailed Feedback} can match or exceed the image-based \textit{Baseline} (see \autoref{fig:6-baseline-vs-detailed-errors}), mirroring findings from ET-PLAN-BENCH \cite{zhang2024et-plan-bench} and LoTa-Bench \cite{choi2024lotaBench}, where structured or explicit feedback reduces the need for perception or interactive plan repair. However, for the strongest vision-capable models (e.g., Kimi-K5.2, GPT-4o), image-based performance remains substantially higher even compared to \textit{Text-Only} with \textit{Detailed Feedback}, suggesting that visual grounding provides information beyond what corrective feedback alone can supply.

\benchmark places substantial demands on visually grounded interactive planning. Models perform markedly better when they can interpret visual observations during execution, and performance degrades when visual grounding or feedback needed for plan repair is removed.

\begin{table}[H]
\centering
\small
\renewcommand{\arraystretch}{1.5}
\begin{tabularx}{\textwidth}{|X|X|X|}
\hline
\multicolumn{1}{|c|}{\normalsize\textbf{No Feedback}} & \multicolumn{1}{c|}{\normalsize\textbf{Success / Failure (Baseline)}} & \multicolumn{1}{c|}{\normalsize\textbf{Detailed Feedback}} \\
\hline

\vspace{-6pt}
\begin{enumerate}[nosep, leftmargin=*]
    \item FIND Plate
    \item FIND Pan
    \item PICKUP Egg
    \item PICKUP Mug
    \item CLEAN Mug
\end{enumerate}
\vspace{-6pt}
&
\vspace{-6pt}
\begin{enumerate}[nosep, leftmargin=*]
    \item FIND Plate Success
    \item FIND Pan Success
    \item PICKUP Egg \textbf{Failure}
    \item PICKUP Mug Success
    \item CLEAN Mug \textbf{Failure}
\end{enumerate}
\vspace{-6pt}
&
\vspace{-6pt}
\begin{enumerate}[nosep, leftmargin=*]
    \item FIND Plate Success
    \item FIND Pan Success
    \item PICKUP Egg \textbf{Failure: Cannot pick up Egg as it is not visible.}
    \item PICKUP Mug Success
    \item CLEAN Mug \textbf{Failure: Mug must be in the SinkBasin to clean.}
\end{enumerate}
\vspace{-6pt} \\
\hline

\end{tabularx}
\caption{\textbf{Past-actions prompt under different feedback conditions.} The left column shows the prompt when the agent receives \textit{No Feedback}, the center column shows the prompt with \textit{Simple Feedback}, and the right column shows the prompt with \textit{Detailed Feedback}.}
\label{tab:2-comparing-past-action-prompts-different-feedbacks}
\end{table}

\subsection{Failure Analysis}
\label{subsec:failure-analysis}

Though we consider success or failure for a Task in a binary fashion, there is some value in digging deeper into the types of failures. We looked at two aspects: (1) why a turn failed, and (2) why a Task terminated early.

\subsubsection{Turn Failure}
\label{subsubsec:turn-failure}

We group turn failures into four categories:

\begin{enumerate}
    \item \textbf{Invalid Action}: The text provided for the Action was not a valid Action
    \item \textbf{Invalid Object}: The text provided for the Object was not a valid Object
    \item \textbf{Undoable}: The Action and Object are valid but the Action could not be performed
    \item \textbf{Unparsable}: The Agent response was malformed
\end{enumerate}

\begin{figure}
    \centering
    \includegraphics[width=0.8\linewidth]{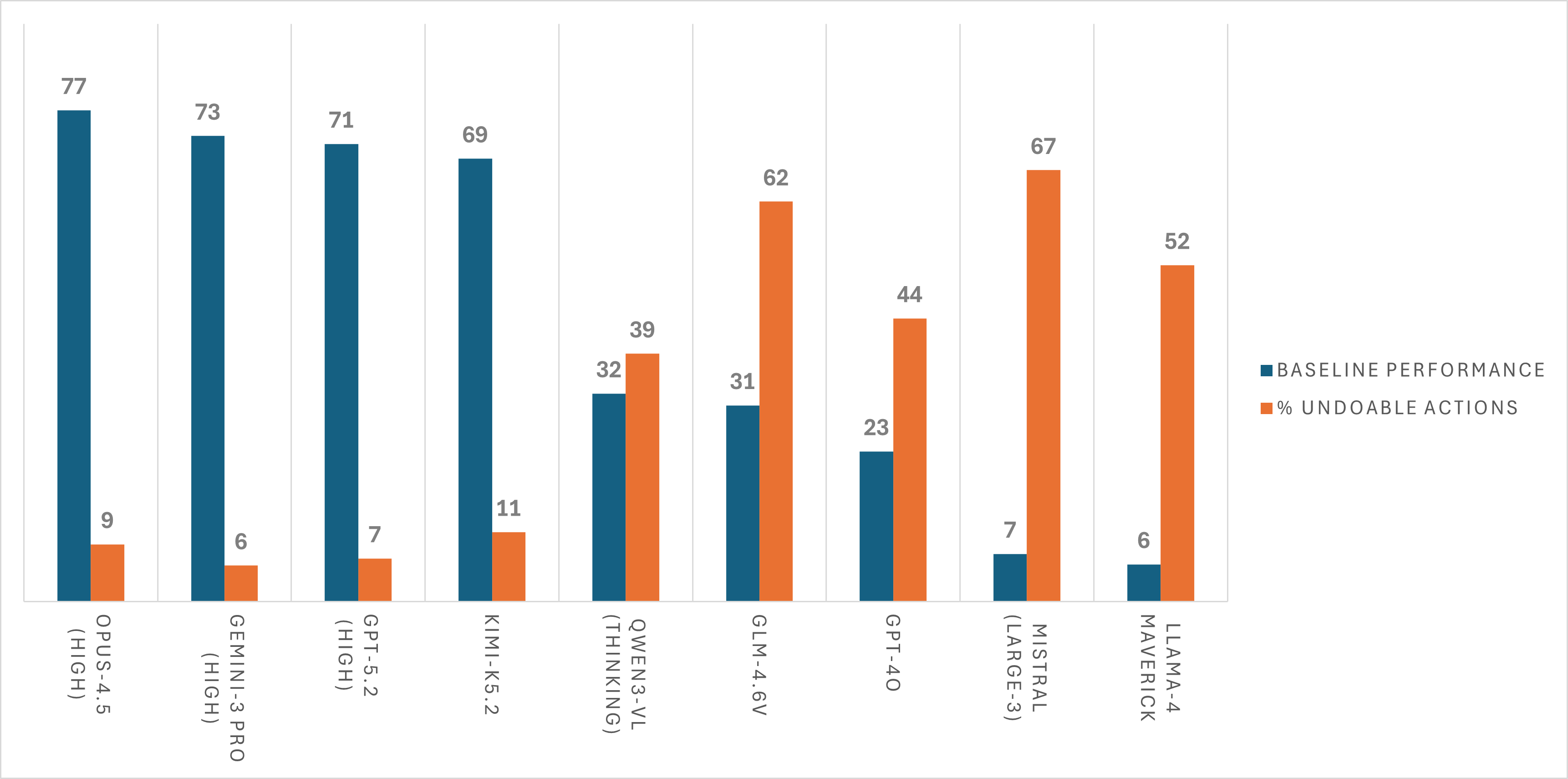}
    \caption{\textbf{Model performance (blue) and percentage of undoable actions (orange).}
Higher performing models produce fewer \textit{undoable} actions, while weaker models generate a larger fraction of actions that cannot be executed. This alignment between success rate and \textit{undoable} action frequency suggests how difficulties in state tracking and plan adjustment contribute directly to overall task failure.}
    \label{fig:7-model-actions-couldnt-be-performed}
\end{figure}

None of the models tested returned \textit{Invalid Actions}. Both \textit{Invalid Object} and \textit{Unparsable} errors were negligible, with only Qwen exhibiting <0.1\% \textit{Invalid Object} and Llama-4 exhibiting <0.1\% \textit{Unparsable}. The percentage of \textit{Undoable} actions aligned closely with overall model performance, with the highest-performing models producing fewer \textit{Undoable} actions (see \autoref{fig:7-model-actions-couldnt-be-performed}).

\subsubsection{Termination Reasons}
\label{subsubsec:termination-reasons}

Looking at the termination conditions for each Task provides some insight into the types of failures. Tasks can terminate in one of four ways:

\begin{enumerate}
    \item \textbf{Success}: Successful completion of the task
    \item \textbf{Max Repeats}: Reached limit for repeating same action or series of actions
    \item \textbf{Max Steps}: Reached maximum number of allowed steps
    \item \textbf{Max Failure}: Reached maximum number of consecutive failures in a row
\end{enumerate}

\begin{figure}
    \centering
    \includegraphics[width=0.8\linewidth]{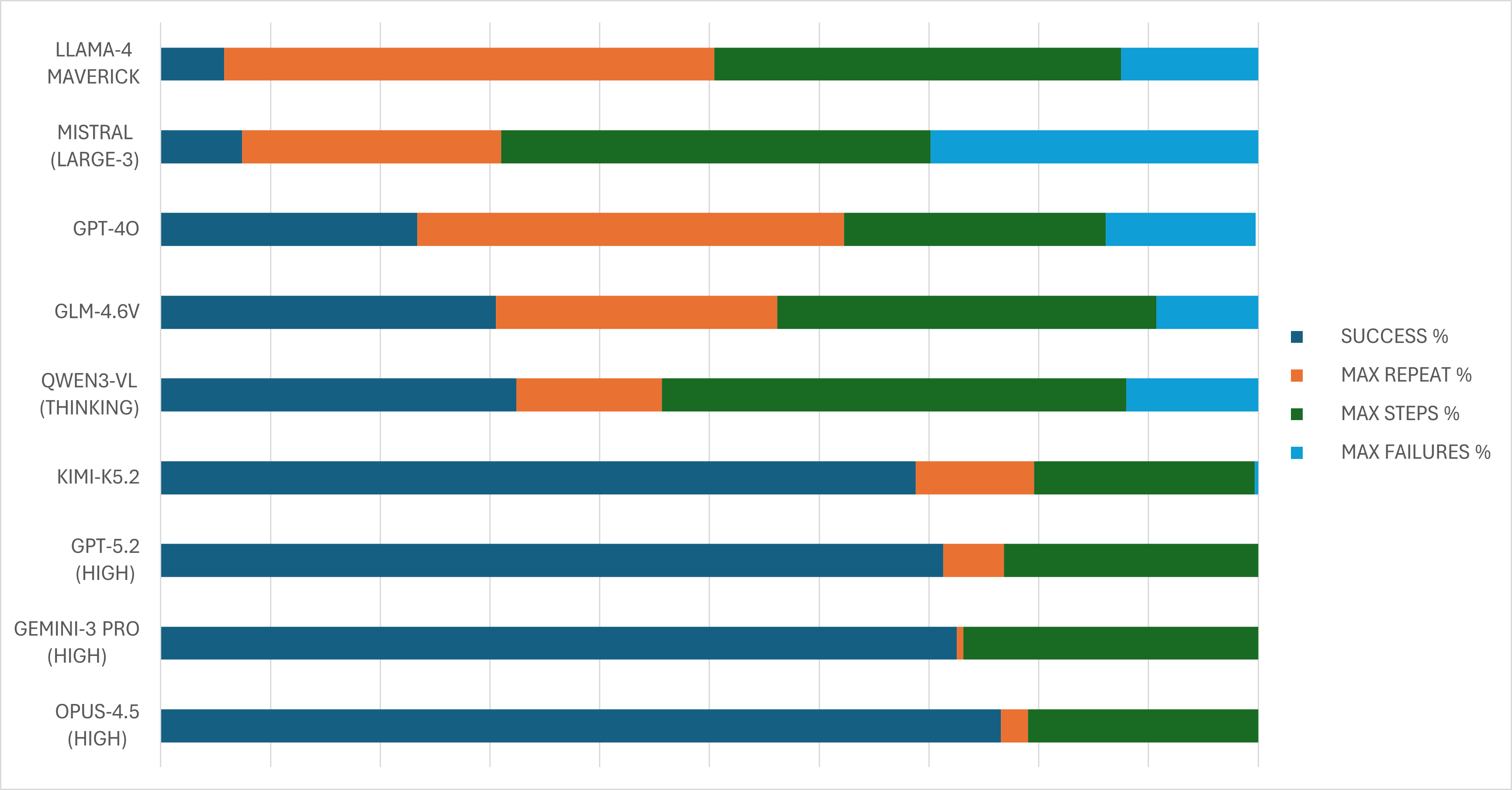}
    \caption{\textbf{Termination reasons.}
Less capable models frequently terminate due to repeated actions/action sequences or consecutive failures, indicating that they become stuck in narrow behavior loops. More capable models terminate primarily due to step limits, reflecting longer but still unsuccessful attempts at plan execution.}
    \label{fig:8-termination-reasons}
\end{figure}

\autoref{fig:8-termination-reasons} shows that the less performant models are more likely to (1) get stuck repeating the same action or short action loop, and (2) repeatedly generate \textit{undoable} actions, leading to early termination.

\subsubsection{Image Conflations}
\label{subsubsec:image-conflations}

\begin{figure}[H]
\centering
\small
\renewcommand{\arraystretch}{1.5}
    \begin{tabularx}{\textwidth}{|>{\centering\arraybackslash}m{0.2\textwidth}|>{\arraybackslash}X|}
        \hline
        \multicolumn{1}{|c|}{\normalsize\textbf{Observation}} & \multicolumn{1}{c|}{\normalsize\textbf{Model Output}} \\
        \hline

        \adjustimage{width=\linewidth, margin={0pt 6pt 0pt 6pt}, valign=m}{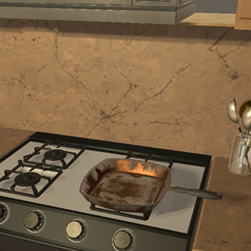} &
        \vspace{-0.78em}\textit{From the image, I can see that the Pan is already on the StoveBurner, and the StoveBurner is turned on because the Pan is visibly on fire. This suggests that the Pan may already be hot, and I need to proceed carefully.} \\
        \hline

        \adjustimage{width=\linewidth, margin={0pt 6pt 0pt 6pt}, valign=m}{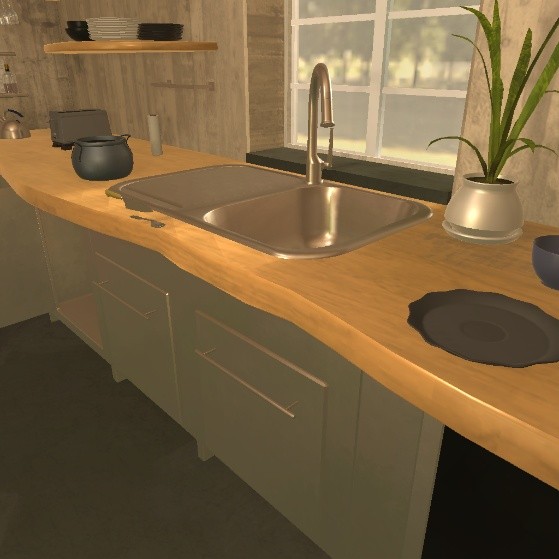} &
        \vspace{0.5em}\textit{Looking at the scene, I see the Mug in the SinkBasin, which appears clean.} \\
        \hline

    \end{tabularx}
    \caption{\textbf{Examples of visual misinterpretations made by GPT-4o.}
    Models occasionally confuse reflections, shadows, or clutter for task-relevant objects, leading to incorrect state inferences and subsequent planning errors.}
    \label{fig:9-image-conflations}
\end{figure}

We also performed an ad-hoc inspection of the Agent failure cases and found that some of the errors stem from visual misinterpretation and not planning. For example, confusing reflections for flames or a Mug (see \autoref{fig:9-image-conflations}). Additional visual errors can be found in \ref{apdx:sample-visual-errors}.

\section{Ablations}
\label{sec:ablations}

In addition to variation in the amount of feedback and \textit{Text-Only} variations, we experimented with the following ablations: (1) removing the hand overlay, (2) removing the \textit{Things to Remember} section, and (3) showing only the current-state image.  These ablations probe different components of visually grounded interactive planning, including state estimation, memory maintenance, and the use of visual history for plan repair.

\subsection{No Hand}
\label{subsec:no-hand}

\begin{figure}[H]
    \centering

    \setlength{\tabcolsep}{5pt}
    \renewcommand{\arraystretch}{1}

    \begin{tabularx}{\textwidth}{X X}
    {\centering\includegraphics[width=0.5\linewidth]{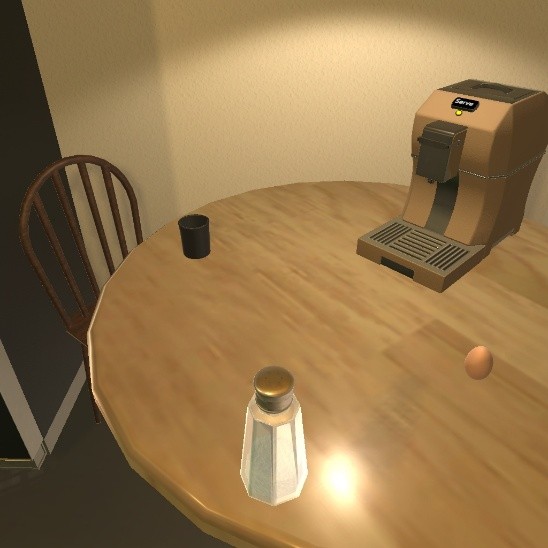}\par} & {\centering\includegraphics[width=0.5\linewidth]{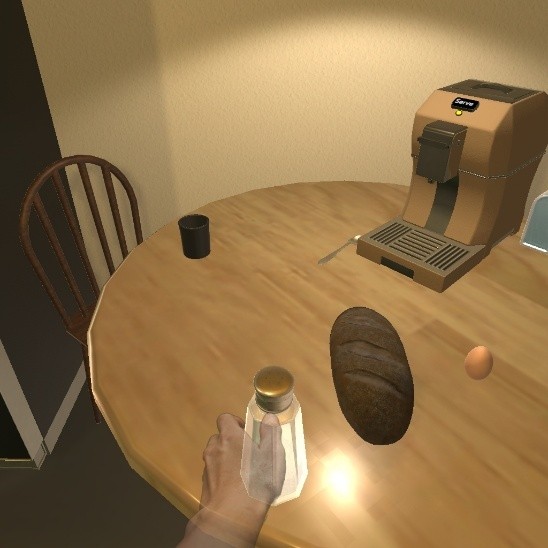}\par} \\

    \end{tabularx}
    \caption{\textbf{Ambiguity in interpreting held objects without the hand overlay.}
In both images, the saltshaker is being held by the agent, but without the superimposed hand overlay it can easily be misinterpreted as resting on the table. This illustrates how subtle spatial cues in AI2-THOR can lead to incorrect state inferences, motivating the inclusion of the hand overlay in \benchmark.}
    \label{fig:10-with-and-without-hand}
\end{figure}

In AI2-THOR, when items are picked up by the Agent, they are displayed in the bottom center of the image without any indication of a hand. In some cases, even for a human, it can be difficult to tell if the object is resting on a counter in front of the agent or being held by it (see \autoref{fig:10-with-and-without-hand}). For this reason, we superimposed a translucent hand over the image whenever an object was being held, to make the held-object state visually unambiguous. The position of the hand was shifted to roughly match the position of the held object on the screen.

Our expectation was that this would help the models disambiguate as the presence of the hand better aligns with the models' training data. We chose translucent rather than opaque because the object visibility check in AI2-THOR cannot account for the superimposed hand image, and we needed to ensure that items are indeed visible to the Agent.

To test our expectations, we evaluated models without the superimposed hand. For these evaluations, we augmented the prompt with additional instructions that explained that held objects would be seen as floating.

\begin{figure}[t]
    \centering
    \begin{subfigure}{\linewidth}
    \caption{No hand overlay}
    \begin{Verbatim}[breaklines,fontsize=\small]
INVENTORY CHECK:
  - Determine if you are holding something and what the OBJECT is
  - Use your VISUAL OBSERVATIONS and ACTION HISTORY FEEDBACK
  - If you are holding an object it will be floating at the bottom center of the current (second) image
  - Your hand will not be visible even if you are holding an object
  - If you see NO floating object in the bottom center, you are NOT holding anything
  - CAUTION: Objects resting on surfaces near the bottom of the frame can be mistaken for held objects.
    - When the visual is ambiguous use ACTION HISTORY FEEDBACK
    \end{Verbatim}
    \end{subfigure}

    \vspace{2em}

    \begin{subfigure}{\linewidth}
    \caption{Hand overlay enabled}
    \begin{Verbatim}[breaklines,fontsize=\small]
INVENTORY CHECK:
  - Determine if you are holding something and what the OBJECT is
  - Use your VISUAL OBSERVATIONS and ACTION HISTORY FEEDBACK
  - If you see an object in your hand or fingers, you ARE holding it (state what it is)
  - If you are holding something then your hand or fingers and the object will be visible at the bottom center of the current (second) image
  - If you see NO hand/fingers or object at the bottom center, you are NOT holding anything
  - CAUTION: Objects resting on surfaces near the bottom of the frame can be mistaken for held objects.
    - When the visual is ambiguous use ACTION HISTORY FEEDBACK
    \end{Verbatim}
    \end{subfigure}

    \vspace{1em}

    \caption{Inventory check prompt instructions with and without hand overlay.}
    \label{fig:hand-overlay-prompt}
\end{figure}

\begin{figure}
    \centering
    \includegraphics[width=0.8\linewidth]{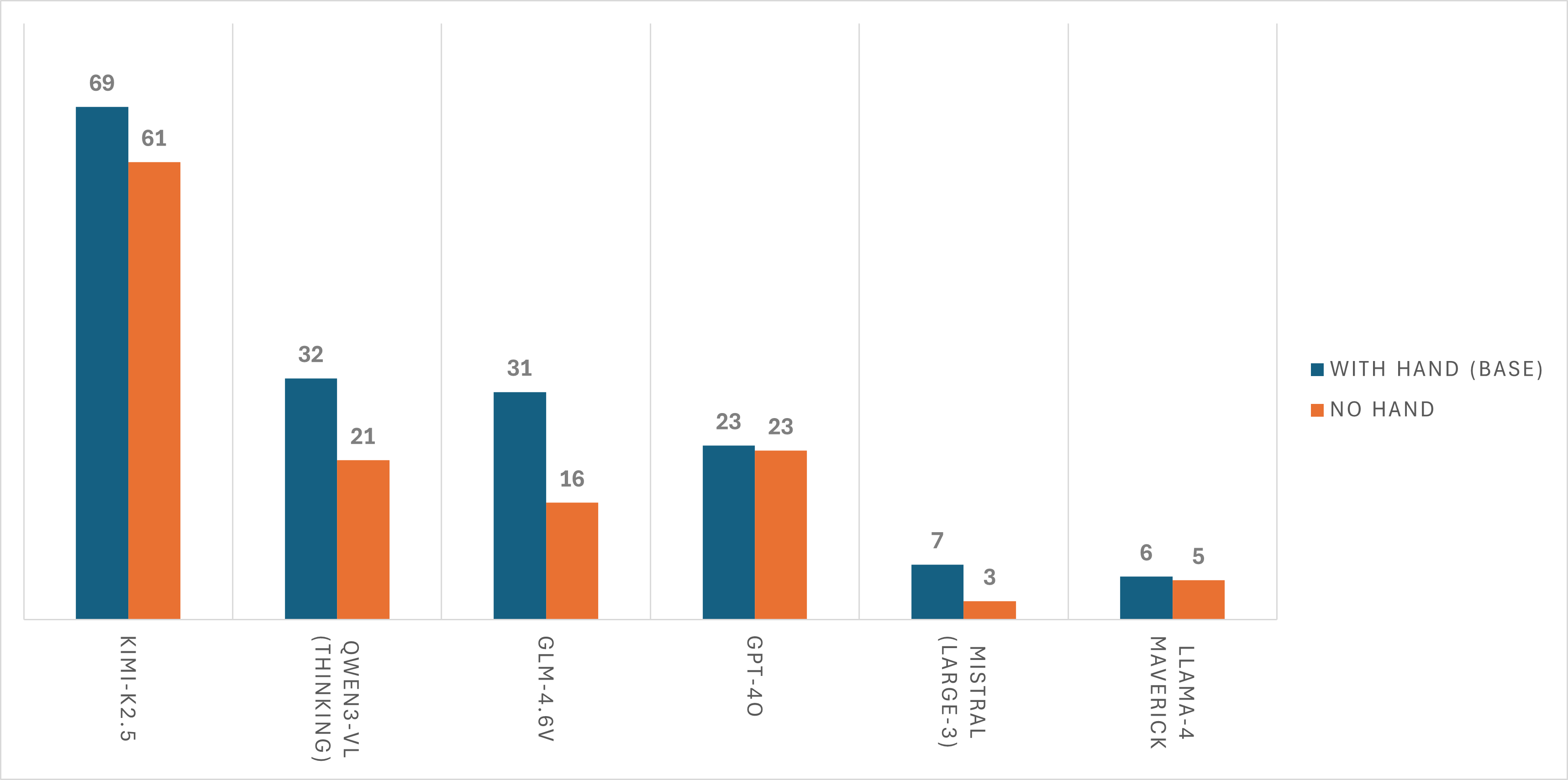}
    \caption{\textbf{Effect of removing the hand overlay on model performance.}
The blue bars show baseline performance with the hand overlay included, while the orange bars show performance when the overlay is removed. The drop in accuracy highlights how models rely on the hand cue to correctly infer when an object is being held.}
    \label{fig:11-performance-with-hand-and-without}
\end{figure}

\begin{figure}[H]
    \centering
    \small
    \begin{tabularx}{\textwidth}{|>{\centering\arraybackslash}p{0.2\textwidth}|>{\arraybackslash}X|}
        \hline
        \multicolumn{1}{|c|}{\normalsize\textbf{Observation}} & \multicolumn{1}{c|}{\normalsize\textbf{Model Output}} \\
        \hline

        \adjustimage{width=\linewidth, margin={0pt 6pt 0pt 6pt}, valign=t}{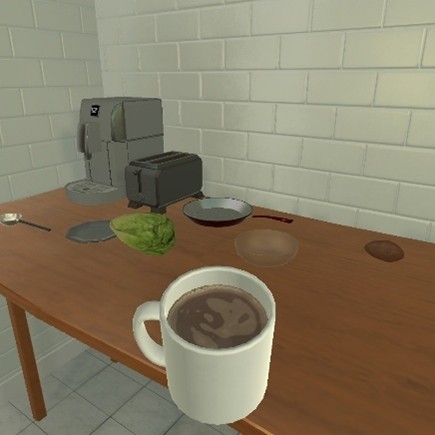} &
        \vspace{0em}\textit{<think>\newline\newline
    First, I check the bottom center of the image for any floating object --- I see nothing floating, so I am NOT holding anything.\newline\newline
    I see a CoffeeMachine, Toaster, Mug (with coffee) and a green object (likely a Lettuce or another vegetable) on the DiningTable...} \\
        \hline

    \end{tabularx}
    \caption{\textbf{Example of a held object being misinterpreted as resting on the table.}
Model incorrectly infers that the coffee mug is on the table, even though it is being held by the agent. This illustrates how the absence of clear hand cues can lead to incorrect state estimation in cluttered scenes.}
    \label{fig:12-model-thinks-coffee-is-on-table}
\end{figure}

All models were found to perform worse without the superimposed hand (see \autoref{fig:11-performance-with-hand-and-without}). We also examined the logs and confirmed that models were having difficulty interpreting images without the hand.  \autoref{fig:12-model-thinks-coffee-is-on-table} shows an example where without the hand overlay the model interpreted a held coffee mug as being on the DiningTable.  This ablation highlights that accurate state estimation, specifically determining whether an object is held or placed, is a prerequisite for correct interactive plan adaptation.

\subsection{Things to Remember}
\label{subsec:things-to-remember}

\begin{figure}[H]
    \centering
    \small
    \renewcommand{\arraystretch}{1}
    \begin{tabularx}{\textwidth}{|>{\centering\arraybackslash}m{0.2\textwidth}|>{\arraybackslash}X|}
        \hline
        \multicolumn{1}{|c|}{\normalsize\textbf{Observation}} & \multicolumn{1}{c|}{\normalsize\textbf{Things to Remember}} \\
        \hline

        \adjustimage{width=\linewidth, margin={0pt 6pt 0pt 6pt}, valign=t}{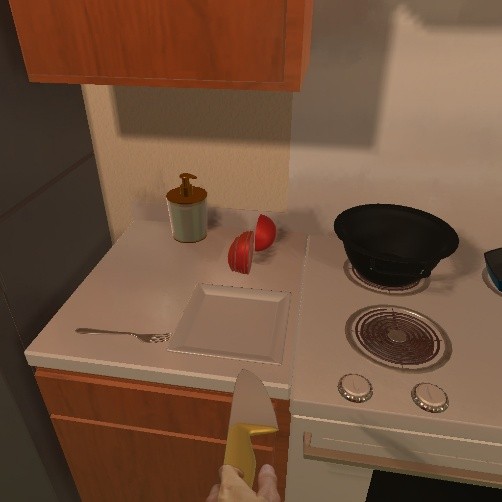} &
        \vspace{0em}
        \textbf{(Claude)}
        \vspace{1em}
        \begin{itemize}[nosep, leftmargin=*]
            \item Completed: Found Knife \checkmark, Picked up Knife \checkmark, Sliced Tomato \checkmark
            \item Current: Put down Knife to free hand
            \item Remaining: Find Plate, Put TomatoSliced on Plate
            \item TomatoSliced is on CounterTop next to SinkBasin
            \item Plate location unknown --- likely in Cabinet
        \end{itemize} \\
        \hline

        \adjustimage{width=\linewidth, margin={0pt 6pt 0pt 6pt}, valign=t}{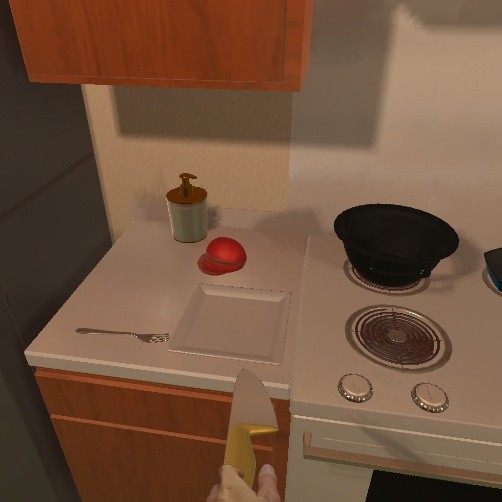} &
        \vspace{0em}
        \textbf{(Qwen)}
        \vspace{1em}
        \begin{itemize}[nosep, leftmargin=*]
            \item Completed: Sliced Tomato \checkmark
            \item Current: Put down Knife to pick up TomatoSliced
            \item Remaining: Pick up TomatoSliced, Put TomatoSliced in Plate
            \item Constraint: Must put down Knife before picking up TomatoSliced (one item at a time)
        \end{itemize} \\
        \hline

        \adjustimage{width=\linewidth, margin={0pt 6pt 0pt 6pt}, valign=t}{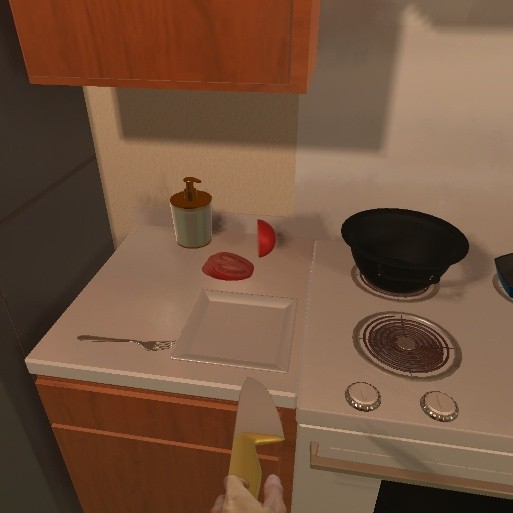} &
        \vspace{0em}
        \textbf{(Llama)}
        \vspace{1em}
        \begin{itemize}[nosep, leftmargin=*]
            \item Completed: Sliced Tomato \checkmark
            \item Current: Find TomatoSliced
            \item Remaining: Put TomatoSliced on Plate
        \end{itemize} \\
        \hline

    \end{tabularx}
    \caption{\textbf{\textit{Things to Remember} prompts for different models.}
The top, middle, and bottom rows show the memory-scaffold prompts used for Claude, Qwen, and Llama, respectively. These examples illustrate how each model is provided with a structured summary of past actions and observations, enabling it to maintain context across steps.}
    \label{fig:13-things-to-remember-comparison}
\end{figure}

During evaluation, Agents are only provided images from the current and previous turn and past actions. We expected that they might struggle without additional context. For example, if the Agent notes that a plate is dirty but then turns away to pick up a sponge, the knowledge of the dirty plate is lost. To mitigate this, we prompted the model to generate a list of \textit{Things to Remember} that is shared with the Agent on the subsequent turn (see the Prompt Example section in \ref{apdx:prompt-example}). Observationally, the content generated by the models for \textit{Things to Remember} was of high quality (see \autoref{fig:13-things-to-remember-comparison}).

\begin{figure}
    \centering
    \includegraphics[width=0.8\linewidth]{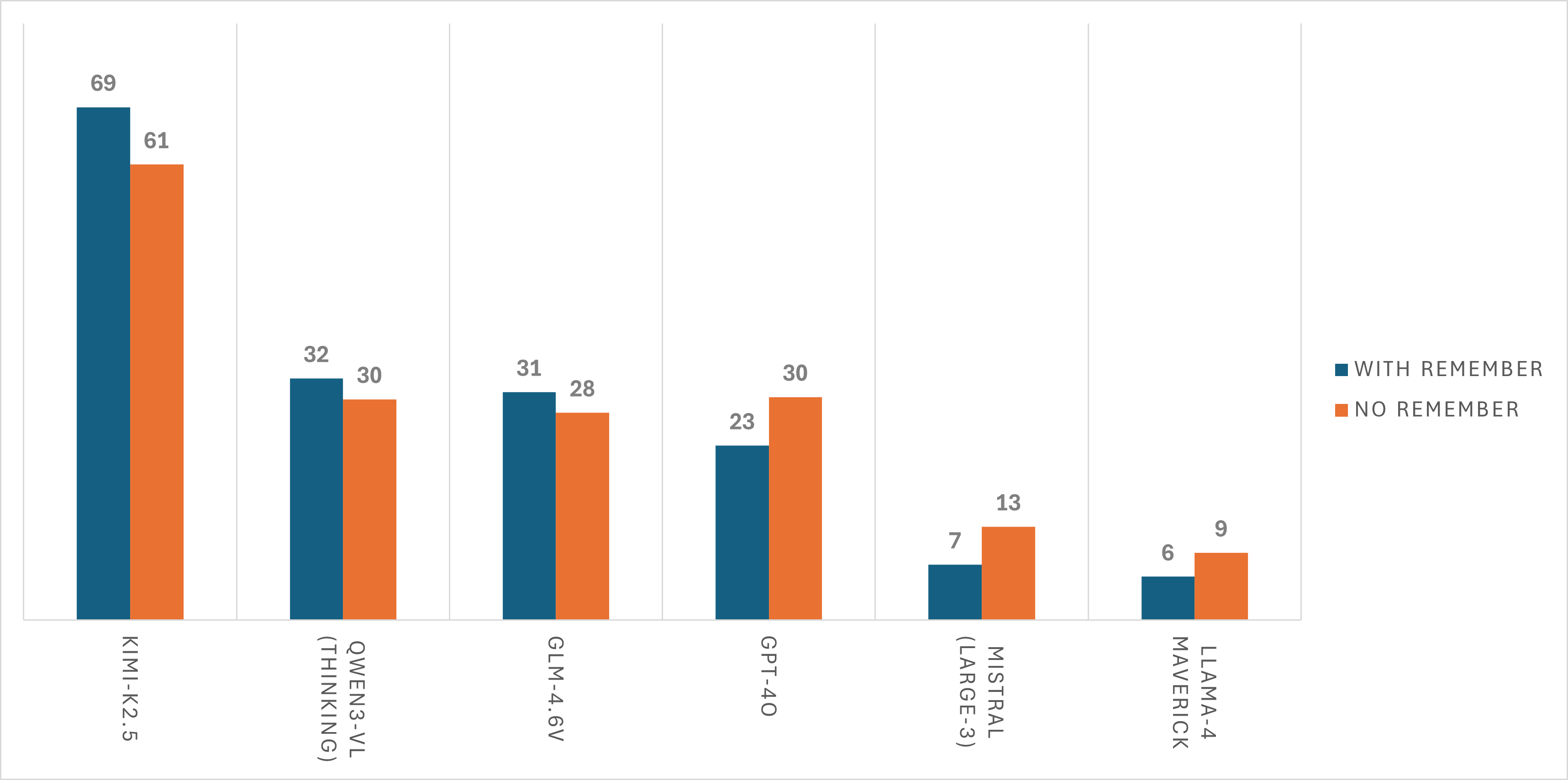}
    \caption{\textbf{Baseline performance (blue) compared to performance without the \textit{Things to Remember} scaffold (orange).}
Removing the memory scaffold produced mixed results: more capable models appeared to benefit from the additional information, while less capable models performed worse without it.}
    \label{fig:14-performance-comparison-memory}
\end{figure}

In the \textit{No Things to Remember} condition, we removed the auxiliary memory summary generated by the model. This ablation tests whether models can maintain task-relevant state purely from the action history and visual observations, without relying on an explicit textual recap.  Removing \textit{Things to Remember} had mixed results. On the surface, the more performant models appeared to benefit from the additional information, while the less performant models suffered (see \autoref{fig:14-performance-comparison-memory}). Because the differences are inconsistent across models, we can't draw a strong conclusions, but the results suggest that explicit memory scaffolding can support interactive planning by helping models maintain task relevant state across steps.

\subsection{Current Image}
\label{subsec:current-image}

In our default setup, we provided the model with two images: the state of the world before the previous action was applied, and the state after. We hypothesized that this temporal context would make it easier to determine the success or failure of each action.

\begin{figure}[t]
    \centering
    \begin{subfigure}{\linewidth}
    \caption{Dual image (baseline)}
    \begin{Verbatim}[breaklines,fontsize=\small]
------------------------------------------------------------
SECTION: IMAGE GUIDANCE
------------------------------------------------------------
- The first image shows the state of the world before the last action was attempted.
- The second image shows the current state of the world (after the last action was attempted).
- Base all your scene observations on the SECOND image (current state).
    \end{Verbatim}
    \end{subfigure}

    \vspace{2em}

    \begin{subfigure}{\linewidth}
    \caption{Single image only}
    \begin{Verbatim}[breaklines,fontsize=\small]
------------------------------------------------------------
SECTION: IMAGE GUIDANCE
------------------------------------------------------------
- Base your scene observations on the input image, which shows the current state of the world.
    \end{Verbatim}
    \end{subfigure}

    \vspace{1em}

    \caption{Image guidance prompt with dual-image baseline vs.\ single current-state image.}
    \label{fig:image-guidance-prompt}
\end{figure}

\begin{figure}
    \centering
    \includegraphics[width=0.8\linewidth]{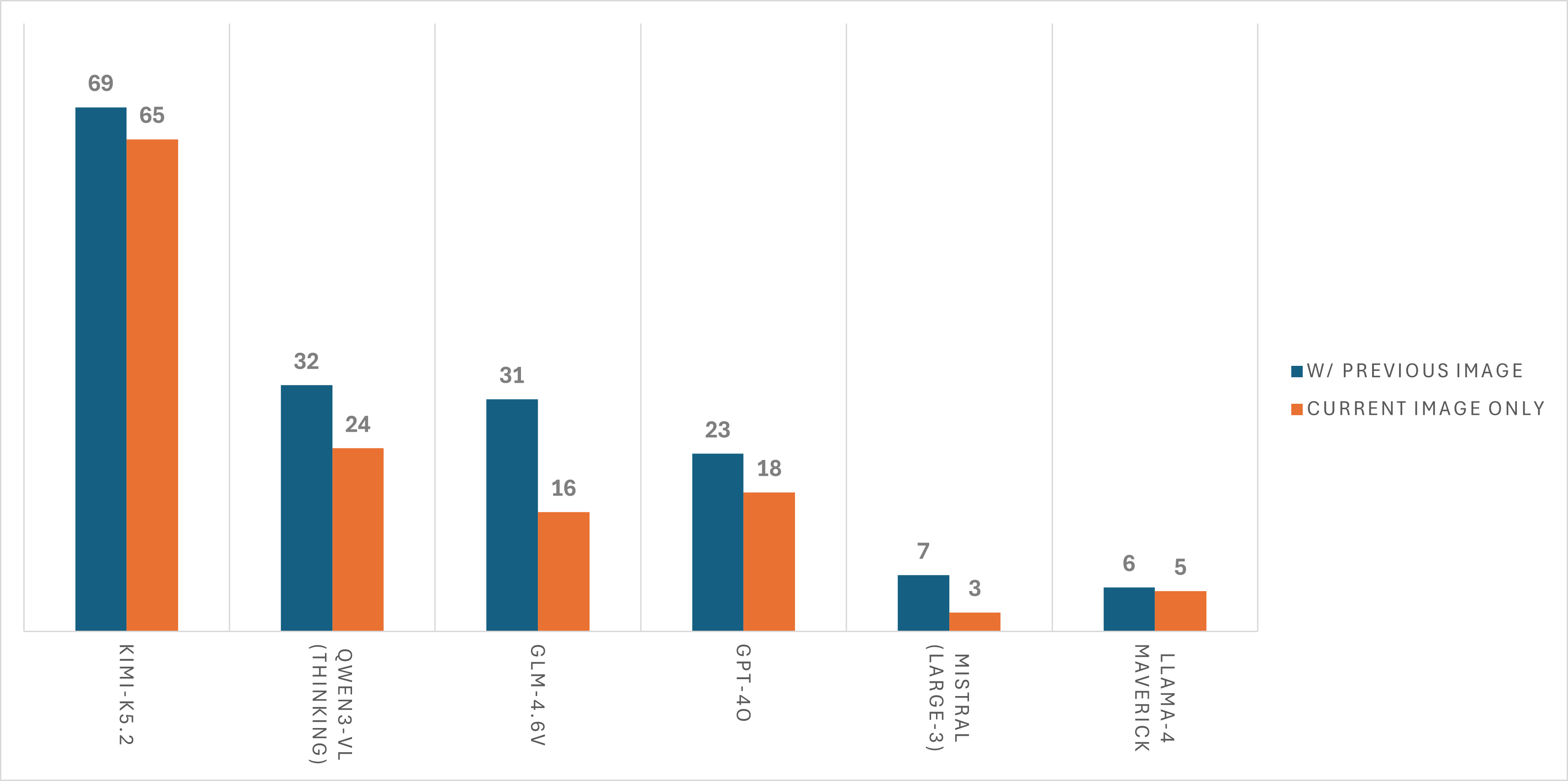}
    \caption{\textbf{Effect of providing previous-state images on model performance. }
The blue bars show performance when the agent is given both the current visual observation and the previous visual observation. The orange bars show performance when only the current observation image is provided. This comparison highlights how access to a short visual history can influence model behavior.}
    \label{fig:15-performance-comparison-prev-image}
\end{figure}

 As shown in \autoref{fig:15-performance-comparison-prev-image}, model performance degrades when only the current state is provided, supporting this hypothesis. We speculate that the previous image helps because (1) the model can compare states to determine the outcome of its actions, and (2) it provides additional spatial context when the viewing direction changes. Short-horizon visual history appears to support interactive planning by helping models detect state changes and repair their plans accordingly.

\section{Discussion}
\label{sec:discussion}

The \benchmark results highlight several consistent challenges for current multimodal agents. Even the strongest models show substantial performance drops when visual input is removed, indicating that the benchmark successfully enforces reliance on multimodal perception rather than letting models fall back on memorized semantic patterns or simulator feedback. This stands in contrast to many existing embodied benchmarks, where text only agents can perform surprisingly well because they receive informative error messages or can rely on stereotyped action templates \cite{ALFRED20, ALFWorld20}.

Across Task types, agents struggle when execution depends on fine grained visual cues. Small differences, like whether a dish is clean or whether a container is open, regularly lead to incorrect state inferences. Visual misinterpretations and clutter mistaken for task relevant objects further show that current vision pipelines remain brittle in cluttered 3D scenes. Once a model adopts an incorrect assumption, it often repeats \textit{undoable} actions or becomes stuck in narrow behavior loops, suggesting state tracking, mismatch detection, and plan adjustment remain significant bottlenecks for interactive planning.

Maintaining an accurate internal model of the environment is another recurring difficulty. Because \benchmark does not provide structured symbolic state, agents must infer and remember object states from images and action histories. Even frontier models frequently lose track of whether an object is clean, where an item was last placed, or whether they are holding something. Adding the \textit{Things to Remember} mechanism improved consistency for some models but did not provide uniform benefits, highlighting that such ad-hoc memory scaffolding is insufficient for reliable state maintenance.  Maintaining task relevant state across interactive execution remains difficult, and prompt-based memory scaffolding provides only inconsistent support. The guidance we provided for generating the memory summary was deliberately simple; a more structured or task-aware memory mechanism might yield more reliable state tracking.

The ablation studies reveal how dependent current systems are on auxiliary cues. Removing the hand overlay or the previous-state image consistently reduced performance, and removing even minimal action feedback made tasks considerably harder. Conversely, providing detailed failure messages sharply improved accuracy, even for \textit{Text-Only} agents. Together, these results show that models rely heavily on cues that simplify state estimation or plan repair, and struggle when required to infer state changes directly from visual observations. This is consistent with the benchmark's goal of evaluating interactive planning under minimal feedback, where progress must come from reasoning grounded in multimodal perception.

\subsection{Benchmark Limitations}
\label{subsec:benchmark-limitations}

While \benchmark is effective as a diagnostic tool, it has limitations. First, the benchmark omits navigation, grasp physics, and other low-level control elements. This abstraction isolates high-level reasoning but reduces ecological validity relative to full embodied settings. Second, the tasks and scenes are confined to AI2-THOR and a curated object set, which limits visual and physical diversity. Third, images are captured under relatively controlled conditions; performance in more varied or photorealistic environments may differ. Fourth, requiring models to produce a full action sequence each turn (even though only the first action is executed) may favor planning-oriented language models over agents designed for reactive, single-step policies. Finally, while minimal feedback is essential for the benchmark's goals, real systems may legitimately incorporate richer sensory inputs, which \benchmark does not capture.  These limitations reflect the benchmark's focus on isolating a specific capability, visually grounded interactive planning.

Overall, \benchmark should be viewed as a targeted benchmark that stresses specific capabilities: visual grounding, conditional reasoning, state tracking, and plan adaptation. It does not aim to represent the full space of embodied intelligence. Instead, it provides a controlled way to expose where current multimodal models break down and where future work is needed, including more robust visual understanding, better internal state representations, and methods for integrating perception and planning more tightly during execution.  This capability focused design is consistent with recent evaluation surveys, which emphasize the need for benchmarks that disentangle high-level reasoning from navigation and low-level control.  \benchmark makes the demands of interactive planning explicit, revealing where current agents still fall short.

\section{Conclusions}
\label{sec:conclusions}

\benchmark was developed to evaluate a capability that is frequently overlooked in existing embodied AI benchmarks: visually grounded interactive planning, the ability of an agent to ground its decisions in visual observations and revise its plans as the environment changes during execution. By removing navigation and low-level manipulation, and by limiting feedback to simple success or failure signals, the benchmark isolates the role of perception in high-level action sequencing. The results across a range of state-of-the-art multimodal models show that this setting remains challenging. Agents struggle with subtle visual distinctions, maintaining coherent internal state, and adjusting plans when initial assumptions prove incorrect. The results show that the current generation of vision-language models struggle in scenarios where perception must drive ongoing decision making and plan repair.  In the terminology of the Embodied AI Evaluation Survey and the Authorea survey, \benchmark directly targets the interactive planning capability category ("what") and evaluates it in a controlled simulator setting ("where"), addressing a gap both surveys identify.

Although \benchmark is intentionally constrained, the framework provides several practical insights. The consistent performance drop in \textit{Text-Only} conditions demonstrates that visual grounding is necessary for success. The ablation studies further show how existing systems may rely on explicit error messages or auxiliary cues. These findings suggest that progress on interactive planning will require models to integrate perception, memory, and planning more tightly and can't treat these components as loosely connected modules.  This echoes the surveys' observation that interactive planning remains under-evaluated and that disentangling high-level reasoning from navigation and low-level control is essential for meaningful capability assessment.

There are also areas where the benchmark can be expanded. Introducing more varied scenes, lighting conditions, and object sets would broaden coverage. Incorporating controlled navigation or limited manipulation elements could help bridge the gap between abstract plan evaluation and full embodied tasks. Additional ways of representing or probing memory, beyond the \textit{Things to Remember} mechanism, may also help clarify how different architectures handle long horizon dependencies. Finally, evaluating models that combine explicit state tracking modules with large scale perceptual front ends could shed light on hybrid approaches that balance learned and structured reasoning.  Such extensions would further align \benchmark with the capability oriented evaluation principles emphasized in recent surveys.

\benchmark offers a targeted, reproducible testbed for studying visually grounded interactive planning. While it does not aim to model all aspects of embodied intelligence, it exposes specific limitations that are likely to matter in more complex environments as well. As multimodal agents continue to advance, benchmarks of this kind will be essential for distinguishing genuine progress in interactive planning from improvements driven by feedback shortcuts or textual priors. By situating \benchmark within a principled evaluation taxonomy, we aim to support more systematic progress on perception-conditioned reasoning and clarify where current models fall short.

\bibliographystyle{unsrt}
\bibliography{references}

@article{yang2025embodiedbench,
  title={EmbodiedBench: Comprehensive Benchmarking Multi-modal Large Language Models for Vision-Driven Embodied Agents},
  author={Yang, Rui and Chen, Hanyang and Zhang, Junyu and Zhao, Mark and Qian, Cheng and Wang, Kangrui and Wang, Qineng and Koripella, Teja Venkat and Movahedi, Marziyeh and Li, Manling and others},
  journal={arXiv preprint arXiv:2502.09560},
  year={2025}
}

@inproceedings{ALFRED20,
  title ={{ALFRED: A Benchmark for Interpreting Grounded
           Instructions for Everyday Tasks}},
  author={Mohit Shridhar and Jesse Thomason and Daniel Gordon and Yonatan Bisk and
          Winson Han and Roozbeh Mottaghi and Luke Zettlemoyer and Dieter Fox},
  booktitle = {The IEEE Conference on Computer Vision and Pattern Recognition (CVPR)},
  year = {2020},
  url  = {https://arxiv.org/abs/1912.01734}
}

@inproceedings{kim2024realfred,
  author    = {Kim, Taewoong and Min, Cheolhong and Kim, Byeonghwi and Kim, Jinyeon and Jeung, Wonje and Choi, Jonghyun},
  title     = {ReALFRED: An Embodied Instruction Following Benchmark in Photo-Realistic Environment},
  booktitle = {ECCV},
  year      = {2024}
}

@inproceedings{ALFWorld20,
  title ={{ALFWorld: Aligning Text and Embodied
           Environments for Interactive Learning}},
  author={Mohit Shridhar and Xingdi Yuan and
          Marc-Alexandre C\^ot\'e and Yonatan Bisk and
          Adam Trischler and Matthew Hausknecht},
  booktitle = {Proceedings of the International Conference on Learning Representations (ICLR)},
  year = {2021},
  url = {https://arxiv.org/abs/2010.03768}
}

@article{li2024behavior1k,
    title   = {BEHAVIOR-1K: A Human-Centered, Embodied AI Benchmark with 1,000 Everyday Activities and Realistic Simulation},
    author  = {Chengshu Li and Ruohan Zhang and Josiah Wong and Cem Gokmen and Sanjana Srivastava and Roberto Martín-Martín and Chen Wang and Gabrael Levine and Wensi Ai and Benjamin Martinez and Hang Yin and Michael Lingelbach and Minjune Hwang and Ayano Hiranaka and Sujay Garlanka and Arman Aydin and Sharon Lee and Jiankai Sun and Mona Anvari and Manasi Sharma and Dhruva Bansal and Samuel Hunter and Kyu-Young Kim and Alan Lou and Caleb R Matthews and Ivan Villa-Renteria and Jerry Huayang Tang and Claire Tang and Fei Xia and Yunzhu Li and Silvio Savarese and Hyowon Gweon and C. Karen Liu and Jiajun Wu and Li Fei-Fei},
    journal = {arXiv preprint arXiv:2403.09227},
    year    = {2024}
}

@inproceedings{anderson2018vln, title={Vision-and-Language Navigation: Interpreting Visually-Grounded Navigation Instructions in Real Environments}, author={Anderson, Peter and Wu, Qi and Teney, Damien and Bruce, Jake and Johnson, Mark and S{\"u}nderhauf, Niko and Reid, Ian and Gould, Stephen and van den Hengel, Anton}, booktitle={Proceedings of the IEEE/CVF Conference on Computer Vision and Pattern Recognition (CVPR)}, pages={3674--3683}, year={2018} }

@article{ai2thor,
  author={Eric Kolve and Roozbeh Mottaghi and Winson Han and
          Eli VanderBilt and Luca Weihs and Alvaro Herrasti and
          Daniel Gordon and Yuke Zhu and Abhinav Gupta and
          Ali Farhadi},
  title={{AI2-THOR: An Interactive 3D Environment for Visual AI}},
  journal={arXiv},
  year={2017}
}

@inproceedings{li2024embodied-agent-interface,
  title={Embodied Agent Interface: Benchmarking LLMs for Embodied Decision Making},
  author={Li, Manling and Zhao, Shiyu and Wang, Qineng and Wang, Kangrui and Zhou, Yu and Srivastava, Sanjana and Gokmen, Cem and Lee, Tony and Li, Li Erran and Zhang, Ruohan and others},
  booktitle={NeurIPS 2024},
  year={2024}
}

@misc{chia2024candodatasetneurosymbolicgrounded,
      title={Can-Do! A Dataset and Neuro-Symbolic Grounded Framework for Embodied Planning with Large Multimodal Models}, 
      author={Yew Ken Chia and Qi Sun and Lidong Bing and Soujanya Poria},
      year={2024},
      eprint={2409.14277},
      archivePrefix={arXiv},
      primaryClass={cs.AI},
      url={https://arxiv.org/abs/2409.14277}, 
}

@article{zhang2024et-plan-bench,
  title={Et-plan-bench: Embodied task-level planning benchmark towards spatial-temporal cognition with foundation models},
  author={Zhang, Lingfeng and Wang, Yuening and Gu, Hongjian and Hamidizadeh, Atia and Zhang, Zhanguang and Liu, Yuecheng and Wang, Yutong and Bravo, David Gamaliel Arcos and Dong, Junyi and Zhou, Shunbo and others},
  journal={arXiv preprint arXiv:2410.14682},
  year={2024}
}

@inproceedings{choi2024lotaBench,
  title={LoTa-Bench: Benchmarking Language-oriented Task Planners for Embodied Agents},
  author={Choi, Jae-Woo and Yoon, Youngwoo and Ong, Hyobin and Kim, Jaehong and Jang, Minsu},
  booktitle={International Conference on Learning Representations (ICLR)},
  year={2024}
}

@misc{su2024actplan1kbenchmark,
      title={ActPlan-1K: Benchmarking the Procedural Planning Ability of Visual Language Models in Household Activities}, 
      author={Ying Su and Zhan Ling and Haochen Shi and Jiayang Cheng and Yauwai Yim and Yangqiu Song},
      year={2024},
      eprint={2410.03907},
      archivePrefix={arXiv},
      primaryClass={cs.CL},
      url={https://arxiv.org/abs/2410.03907}, 
}

@inproceedings{procthor,
  author={Matt Deitke and Eli VanderBilt and Alvaro Herrasti and
          Luca Weihs and Jordi Salvador and Kiana Ehsani and
          Winson Han and Eric Kolve and Ali Farhadi and
          Aniruddha Kembhavi and Roozbeh Mottaghi},
  title={{ProcTHOR: Large-Scale Embodied AI Using Procedural Generation}},
  booktitle={NeurIPS},
  year={2022},
  note={Outstanding Paper Award}
}

@inproceedings{szot2021habitat,
  title     =     {Habitat 2.0: Training Home Assistants to Rearrange their Habitat},
  author    =     {Andrew Szot and Alex Clegg and Eric Undersander and Erik Wijmans and Yili Zhao and John Turner and Noah Maestre and Mustafa Mukadam and Devendra Chaplot and Oleksandr Maksymets and Aaron Gokaslan and Vladimir Vondrus and Sameer Dharur and Franziska Meier and Wojciech Galuba and Angel Chang and Zsolt Kira and Vladlen Koltun and Jitendra Malik and Manolis Savva and Dhruv Batra},
  booktitle =     {Advances in Neural Information Processing Systems (NeurIPS)},
  year      =     {2021}
}

@article{webarena2023,
  title={WebArena: A Realistic Web Environment for Building Autonomous Agents},
  author={Zhou, Shuyan and Xu, Frank F and Zhang, Yilun and Zhang, Tianyi and Li, Xinyun and Wong, Nelson F. and others},
  journal={arXiv preprint arXiv:2307.13854},
  year={2023}
}

@inproceedings{webshop2022,
  title={WebShop: Towards Scalable Real-World Web Interaction with Grounded Language Agents},
  author={Yao, Shunyu and Zhao, Jeffrey and Yu, Dian and Narasimhan, Karthik and others},
  booktitle={NeurIPS},
  year={2022}
}

@article{osworld2023,
  title={OSWorld: Benchmarking Multimodal Agents for Computer Use},
  author={Wang, Yizhong and Chen, Yujia and Liu, Hao and others},
  journal={arXiv preprint arXiv:2312.00752},
  year={2023}
}

@article{authoreaSurvey,
  title   = {Embodied AI: A Comprehensive Survey},
  author  = {Xie, Zhaoyuan and others},
  journal = {Authorea Preprints},
  year    = {2024},
  doi     = {10.22541/au.177023340.02874343},
  url     = {https://www.authorea.com/doi/full/10.22541/au.177023340.02874343}
}

@misc{embodiedaisurvey,
  title        = {Embodied AI Evaluation Survey},
  author       = {Embodied AI Community},
  howpublished = {\url{https://github.com/EmbodiedAISurvey/Embodied-AI-Eval-Survey}},
  year         = {2024},
  note         = {Accessed: 2025-02-20}
}

\clearpage
\appendix
\section{Appendix}
\label{sec:appendix}

\subsection{Action Space}
\label{apdx:action-space}

The agent can perform the following high-level actions:

\begin{itemize}
    \item \textbf{CLEAN \{object\_name\}} -- Clean \textit{\{object\_name\}}.
    \item \textbf{CLOSE \{container\_name\}} -- Close \textit{\{container\_name\}}.
    \item \textbf{EMPTY \{container\_name\}} -- Pour out liquid from \textit{\{container\_name\}}.
    \item \textbf{DRINK \{container\_name\}} -- Drink liquid from \textit{\{container\_name\}}.
    \item \textbf{FIND \{object\_name\}} -- Navigate to  \textit{\{object\_name\}}.
    \item \textbf{OPEN \{container\_name\}} -- Open \textit{\{container\_name\}}.
    \item \textbf{PICKUP \{object\_name\}} -- Pick up \textit{\{object\_name\}} and hold it.
    \item \textbf{PUT \{container\_name\}} -- Place the held item into \textit{\{container\_name\}}.
    \item \textbf{SLICE \{object\_name\}} -- Slice \textit{\{object\_name\}} or crack an egg.
    \begin{itemize}
        \item To "SLICE", the agent must be holding a knife. This action is also used for cracking an egg, in which case no knife needs to be held.
    \end{itemize}
    \item \textbf{SPRAY \{object\_name\}} -- Spray \textit{\{object\_name\}} using the held object.
    \item \textbf{TOGGLE\_ON \{object\_name\}} -- Turn on \textit{\{object\_name\}}.
    \item \textbf{TOGGLE\_OFF \{object\_name\}} -- Turn off \textit{\{object\_name\}}.
\end{itemize}

Most actions work as expected, but two warrant special explanation.

\textbf{FIND} -- abstracts away navigation entirely: the benchmark handles the low-level actions required to locate the target object and positions the agent in front of it. If the object is inside a container, the agent is placed in front of that closed container.

\textbf{PUT} -- handles placement logistics. The simulator verifies that space exists for the held object at the target location. If not, it automatically tries other containers of the same type. For example, if `PUT Cabinet` fails because `Cabinet\_1` is full, it will try `Cabinet\_2`, `Cabinet\_3`, and so on, navigating to each and opening as needed until it finds available space or exhausts all candidates. Note that the initial `PUT` requires the agent to already be facing a valid, visible container (the automatic navigation occurs only when retrying alternative containers).

\subsection{Observation Space}
\label{apdx:observation-space}

At each turn, the agent receives the following inputs:

\textbf{Action history:}
A numbered list of previously attempted actions with simple feedback indicating success or failure (e.g., `PICKUP Mug Success` or `CLEAN Mug Failure`).

\textbf{Suggested plan:}
Rather than outputting only a single next action, the model outputs a complete action sequence covering all remaining steps. However, only the first action in the sequence is executed at each turn. In subsequent turns, the previous plan is provided back to the model alongside the newly observed state, allowing it to re-evaluate and revise the plan before selecting its next action.

\textbf{Images:}
Two 1024$\times$1024 RGB images are provided: the current scene and the state before the previous action. Comparing these images enables the agent to infer whether its last action succeeded.

\textbf{Hand overlay:}
When the agent is holding an object, a translucent hand (60\% opacity) is superimposed on the image at a predefined location aligned with the held item. This reduces ambiguity between held items and objects resting on nearby surfaces.

\subsubsection{Prompt Structure}
\label{apdx:prompt-structure}

Each prompt comprises the following components:

\begin{itemize}
    \item \textbf{High-level goal} -- e.g., "Make a slice of toast and serve it on a plate."
    \item \textbf{Available objects} -- the list of objects present in the current scene.
    \item \textbf{Available actions} -- the action space with syntax and descriptions.
    \item \textbf{World rules} -- constraints governing valid action sequences (e.g., "Items must be placed in the sink before cleaning," "The microwave must be open before placing items inside"). See \ref{apdx:prompt-example} for an example.
    \item \textbf{Agent notes:}
    \begin{itemize}
        \item Action history with success/failure feedback.
    \end{itemize}
    \begin{itemize}
        \item The suggested plan from the previous response.
    \end{itemize}
    \item \textbf{Task instructions} -- detailed guidance on how to structure reasoning and output.
    \item \textbf{Examples} -- three in-context examples demonstrating correct reasoning and action format.
\end{itemize}

For a full prompt example, see \ref{apdx:prompt-example}.

\subsection{Benchmark Generation}
\label{apdx:benchmark-generation}

\benchmark uses a pre-generated, fixed test set to ensure reproducibility and enable fair comparison across models. Each task instance is referred to as a \textit{plan} (a specific configuration of a task type, scene, object placements, and variation conditions). All plans are constructed before any model evaluation occurs, ensuring that every model is tested on the same curated collection of scenarios.

A plan captures everything required to define a solvable task: which scene to use, where objects are placed, which preconditions apply (e.g., whether the plate starts dirty), and the goals that must be fulfilled for the given task. Because object placement is randomized and tasks involve conditional logic (e.g., dirty items must be cleaned before use), the same task type can yield hundreds of distinct plans, each requiring a different sequence of actions to complete.

\benchmark scenes vary in which objects are present, creating natural variation across plans. For each task, the positions of objects are additionally randomized while ensuring that all task-relevant items (e.g., dishes, food items) remain visible and are not spawned inside closed containers such as drawers or refrigerators.

\subsubsection{Object Slicing}
\label{apdx:object-slicing}

\begin{figure}[H]
    \centering
    \small
    \renewcommand{\arraystretch}{1}
    \begin{tabularx}{\textwidth}{|>{\centering\arraybackslash}X|>{\centering\arraybackslash}X|>{\centering\arraybackslash}X|}
        \hline
        \multicolumn{1}{|c|}{\normalsize\textbf{Unsliced}} & \multicolumn{1}{c|}{\normalsize\textbf{Sliced}} & \multicolumn{1}{c|}{\normalsize\textbf{With Nudge}} \\
        \hline

        \adjustimage{width=0.75\linewidth, margin={0pt 6pt 0pt 6pt}, valign=t}{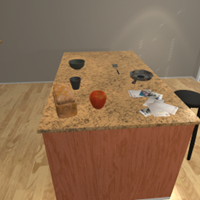} &
        \adjustimage{width=0.75\linewidth, margin={0pt 6pt 0pt 6pt}, valign=t}{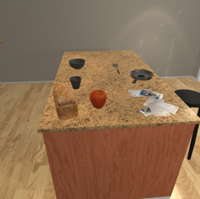} &
        \adjustimage{width=0.75\linewidth, margin={0pt 6pt 0pt 6pt}, valign=t}{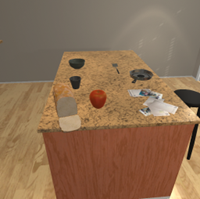} \\
        \hline

        \adjustimage{width=0.75\linewidth, margin={0pt 6pt 0pt 6pt}, valign=t}{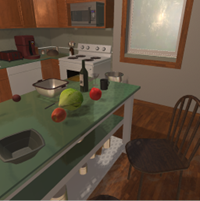} &
        \adjustimage{width=0.75\linewidth, margin={0pt 6pt 0pt 6pt}, valign=t}{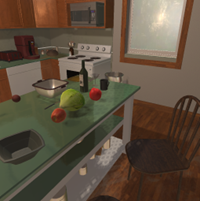} &
        \adjustimage{width=0.75\linewidth, margin={0pt 6pt 0pt 6pt}, valign=t}{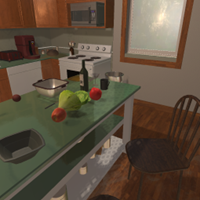} \\
        \hline

    \end{tabularx}
\caption{\textbf{Adding small positional nudges to sliced items to improve visibility. }
Sliced objects in AI2-THOR often overlap or collapse into nearly indistinguishable configurations. Adding slight positional nudges after slicing makes the resulting pieces more visually separable, reducing ambiguity.}
\label{fig:16-adding-nudge-after-slice}
\end{figure}

In AI2-THOR, sliced objects often appear almost identical to their unsliced counterparts. \autoref{fig:16-adding-nudge-after-slice} presents examples of unsliced (left) and sliced (center) bread and lettuce. Models frequently struggled to determine whether an item was sliced. To address this issue, code was added to nudge the sliced items apart, making it easier to identify that they had been sliced. The right-hand image shows the nudged versions of the sliced bread and lettuce.

\subsubsection{Plan Sanitization}
\label{apdx:plan-sanitization}

To maintain consistency and avoid ambiguous or unsolvable states, several modifications are applied during scene preparation, before a given plan is finally included in the final test set:

\begin{itemize}
    \item Remove duplicate items to avoid ambiguous commands (e.g., multiple vases making "PICKUP Vase" unclear).
    \item Avoid placing silverware in the sink, as deep basins can occlude them from the agent's viewpoint.
    \item Remove visually confusable objects, such as the PepperShaker, which is nearly indistinguishable from the SaltShaker in most views.
    \item Remove objects that cannot be stably placed, either because they fall through surfaces or cannot contain other items.
    \item Exclude malformed AI2-THOR scenes, such as kitchen floor plans where objects spawn below the floor or where an invisible mesh blocks placement (e.g., FloorPlans 8, 9, and 17).
\end{itemize}

Not every random configuration yields a solvable task (for example, a required object might spawn inside an unreachable container). To ensure that all plans in the benchmark are completable, each configuration is validated with a deterministic state machine before inclusion in the test set. The following pseudocode describes the plan generation process:

\begin{Verbatim}
FOR each task_type in [cook_egg, make_toast, put_away, ...]:
    FOR each scene in compatible_scenes(task_type):
        FOR each variation in [clean, dirty_plate, dirty_pan, ...]:

            1. Initialize scene with random seed
            2. Remove duplicate objects (keep one per type)
            3. Remove known buggy objects
            4. Place task-relevant objects on visible surfaces
            5. Apply variation conditions (e.g., make Plate dirty)

            6. Attempt to complete task using deterministic state machine:
               - Has perfect knowledge of the scene
               - Determines valid interaction positions
               - Checks object compatibility (e.g., slice fits in bowl)

            7. IF state machine succeeds:
                   Save configuration to benchmark
               ELSE:
                   Discard configuration
\end{Verbatim}

This process guarantees that every accepted plan has at least one valid solution path, while the randomized placements introduce meaningful diversity in the required action sequences. The final benchmark test set consists of 108 plans spanning all 12 task types.

\subsection{Task Completion Criteria}
\label{apdx:task-completion-criteria}

A task is considered complete when all associated goals have been met. The goals fall into the following categories:

\begin{itemize}
    \item \textbf{State Goals}: Object properties that must be in a specific state
    \item \textbf{Location Goals}: Objects that must be placed in or on a specific container type
    \item \textbf{Put-Away Goals}: Objects that must be stored in their designated storage locations
    \item \textbf{Action Goals}: Specific actions that must have been performed (e.g., drink coffee, spray mirror)
    \item \textbf{Contents Goals}: Containers that must contain specific items
\end{itemize}

Goals are evaluated after every agent action is applied to the simulation. If all goals are satisfied, the task is considered successfully completed.

\subsubsection{Task-Specific Goals}
\label{apdx:task-specific-goals}

For all kitchen tasks, the world must be left in a reasonable state; as such, the following goals are shared by all tasks that occur in a kitchen scene:

\begin{table}[H]
\centering
\small
\renewcommand{\arraystretch}{1.5}
\begin{tabularx}{\textwidth}{|c|X|}
\hline
\multicolumn{1}{|c|}{\normalsize\textbf{Scene}} & \multicolumn{1}{c|}{\normalsize\textbf{Goals}} \\
\hline

Kitchen &
\vspace{-6pt}
\begin{enumerate}[nosep, leftmargin=*]
    \item CoffeeMachine must be off
    \item Toaster must be off
    \item StoveBurner must be off
    \item Microwave must be closed
    \item Fridge must be closed
    \item Faucet must be off
\end{enumerate}
\vspace{-6pt} \\
\hline

\end{tabularx}
\caption{\textbf{Kitchen-specific goals shared across all tasks.}
Because every kitchen task requires leaving the environment in a reasonable state, all tasks in kitchen scenes share a common set of cleanup-oriented goals. This table lists those goals, which ensure that agents not only complete the primary task but also restore the scene to an acceptable post-task configuration.}
\label{tab:3-kitchen-goals}
\end{table}

\autoref{tab:task-specific-goals} shows the actual goals for all of the tasks we employ in \benchmark.

\begin{table}[H]
\centering
\small
\renewcommand{\arraystretch}{1.5}
\begin{tabularx}{\textwidth}{|>{\raggedright\arraybackslash}p{4.2cm}|c|>{\raggedright\arraybackslash}X|}
\hline
\multicolumn{1}{|c|}{\normalsize\textbf{Task}} & {\normalsize\textbf{Scene}} & \multicolumn{1}{c|}{\normalsize\textbf{Goals}} \\
\hline

Microwave a potato and serve it in a bowl & Kitchen &
\vspace{-6pt}
\begin{itemize}[nosep, leftmargin=*]
    \item Potato must be cooked
    \item Potato must be in a Bowl
    \item Bowl must be clean if it starts dirty
\end{itemize}
\vspace{-6pt} \\
\hline

Cook an egg in a pan and serve it on a plate & Kitchen &
\vspace{-6pt}
\begin{itemize}[nosep, leftmargin=*]
    \item Egg must be cooked
    \item EggCracked must be on a Plate
    \item Pan must be clean if it starts dirty
    \item Plate must be clean if it starts dirty
\end{itemize}
\vspace{-6pt} \\
\hline

Make a slice of toast and serve it on a plate & Kitchen &
\vspace{-6pt}
\begin{itemize}[nosep, leftmargin=*]
    \item Bread must be cooked/toasted
    \item BreadSliced must be on a Plate
    \item Plate must be clean if it starts dirty
\end{itemize}
\vspace{-6pt} \\
\hline

Fry a potato slice and serve it on a plate & Kitchen &
\vspace{-6pt}
\begin{itemize}[nosep, leftmargin=*]
    \item PotatoSliced must be cooked
    \item PotatoSliced must be on a Plate
    \item Pan must be clean if it starts dirty
    \item Plate must be clean if it starts dirty
\end{itemize}
\vspace{-6pt} \\
\hline

Put away the \{Food\} (e.g., Apple, Bread, Egg, Lettuce, Potato, Tomato) & Kitchen &
\vspace{-6pt}
\begin{itemize}[nosep, leftmargin=*]
    \item Each specified food item must be put away (stored in Fridge)
    \item Items must be clean if they start dirty
\end{itemize}
\vspace{-6pt} \\
\hline

Put away the \{Dishes\} (e.g., Bowl, Cup, Mug, Plate, Pot) & Kitchen &
\vspace{-6pt}
\begin{itemize}[nosep, leftmargin=*]
    \item Each specified dish must be put away (stored in Cabinet)
    \item Items must be clean if they start dirty
\end{itemize}
\vspace{-6pt} \\
\hline

Put away the \{Silverware\} (e.g., ButterKnife, Fork, Knife, Ladle, Spatula, Spoon) & Kitchen &
\vspace{-6pt}
\begin{itemize}[nosep, leftmargin=*]
    \item Each specified silverware item must be put away (stored in Drawer)
    \item Items must be clean if they start dirty
\end{itemize}
\vspace{-6pt} \\
\hline

Slice the \{Lettuce/Apple/Tomato\} and put a piece in the \{Plate/Bowl/Pan/Pot\} & Kitchen &
\vspace{-6pt}
\begin{itemize}[nosep, leftmargin=*]
    \item A slice of the specified item must be in each specified container
    \item Each container must be clean if it starts dirty
\end{itemize}
\vspace{-6pt} \\
\hline

\end{tabularx}
\caption{\textbf{Task-specific goals for each benchmark scenario.}
In addition to the shared kitchen goals listed in \autoref{tab:3-kitchen-goals}, each task includes its own task-specific goals that define the required object states or interactions needed to complete that scenario. This table lists those per-task goals, which capture the unique requirements and success conditions for each individual task.}
\label{tab:task-specific-goals}
\end{table}

\subsection{Undoable Actions}
\label{apdx:undoable-actions}
\begin{figure}
    \centering
    \includegraphics[height=0.45\linewidth]{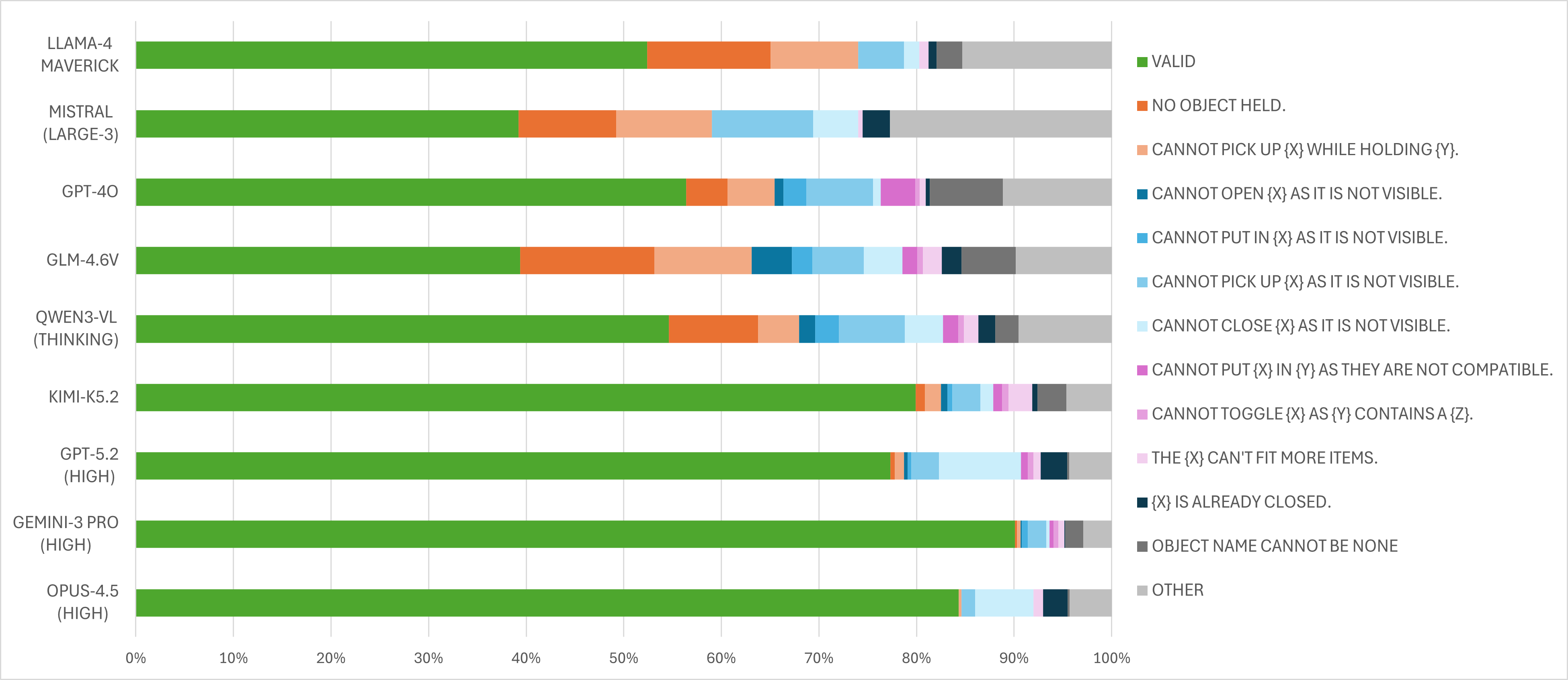}
    \caption{\textbf{Distribution of undoable-action error types across models in \benchmark.}
Each bar shows the proportion of undoable actions attributed to different failure categories (e.g., holding state errors, visibility errors, incompatibility errors). Higher performing models exhibit fewer holding-state errors, while visibility related errors remain relatively consistent across models.}
    \label{fig:top-errors}
\end{figure}

\textit{Undoable} actions occur when the model produces a syntactically valid action and object name, but the environment rules make the action impossible to execute.
\autoref{fig:top-errors} summarizes the most common undoable action categories across models. Lower performing models show a disproportionately high rate of errors related to object holding state (e.g., attempting to pick up an object while already holding another). In contrast, visibility related errors occur at similar rates across both strong and weak models.

\begin{figure}
    \centering
    \includegraphics[height=1.2\linewidth]{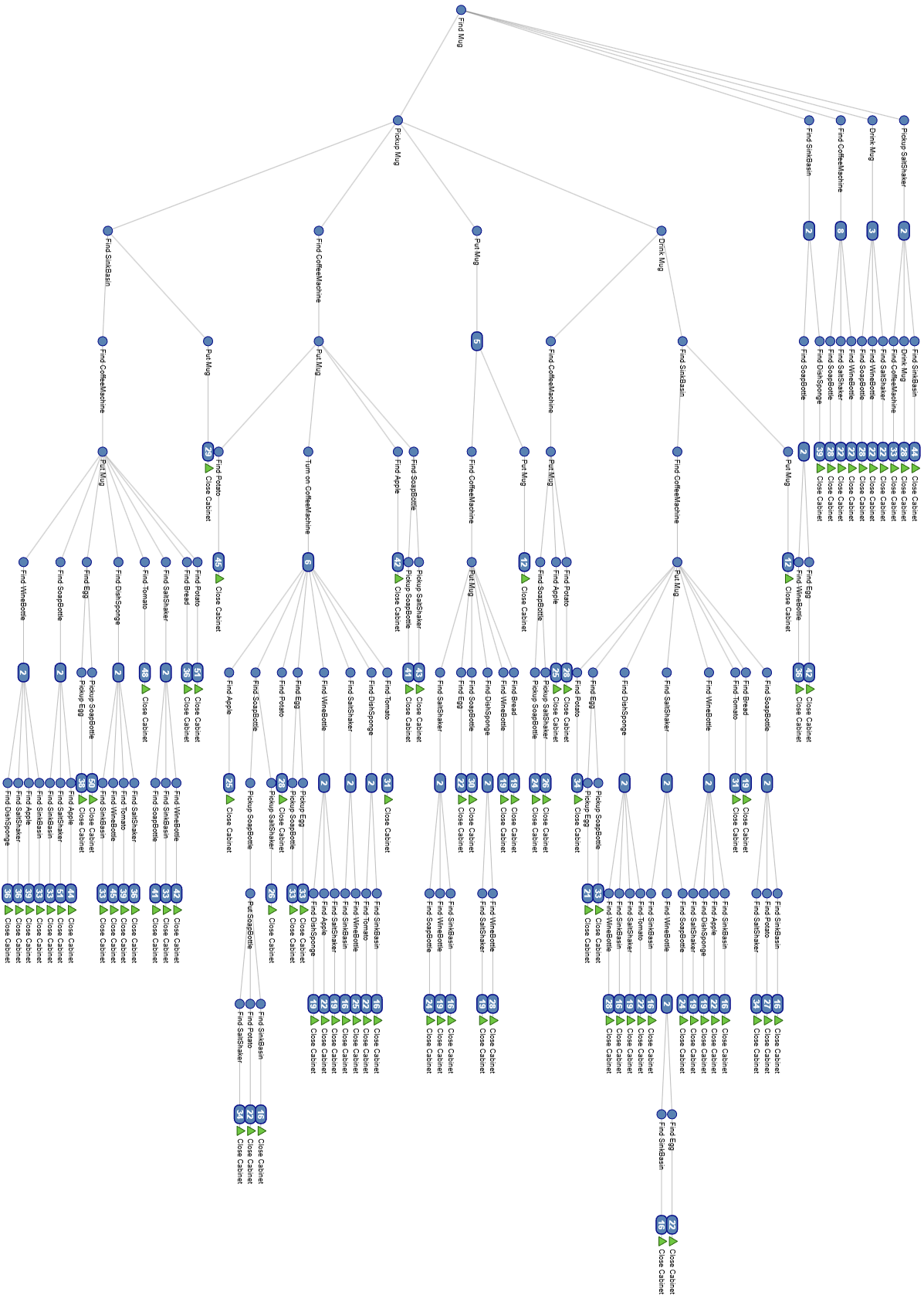}
    \caption{\textbf{Conditional branching from a single instruction in \benchmark.}
Because objects are initialized in different configurations (e.g., clean vs. dirty) and placed in different positions, a single Task instruction requires distinct action plans for each initialization, even within the same Scene. Shown here is the Task \textit{"Consume coffee from a mug, then wash and store the mug"} in "FloorPlan3", where each initial state leads to a different valid execution sequence.}
    \label{fig:coffee-plans}
\end{figure}

\subsection{Reproducibility Notes}
\label{apdx:reproducibility-notes}

This section documents the experimental details needed to reproduce our reported numbers.

\benchmark uses a pre-generated, fixed test set of 108 tasks.
Every evaluation run for a given model uses the same task set; however, even with model
temperature set to zero, we observe non-negligible variance in final
performance scores. We attribute this to inherent stochasticity in the
AI2-THOR simulation as well as residual randomness during model inference.

To mitigate this variance, we evaluated most models over multiple
independent runs and report the \textbf{median} across runs.
Due to budget constraints, \texttt{GPT-5.2} was evaluated with a single run only.
% Per-model run counts are listed in \autoref{tab:model-hyperparams}.

Models were evaluated via two serving routes:
(i)~\textbf{Azure Foundry} and (ii)~\textbf{OpenRouter}.
For Azure Foundry, we used the lowest available content-filter sensitivity; for OpenRouter, we used provider defaults.

\begin{table}[H]
    \centering
    \small
    \renewcommand{\arraystretch}{1.35}
    \begin{tabularx}{\textwidth}{|
    >{\raggedright\arraybackslash}p{5.4cm}|
    >{\raggedright\arraybackslash}X|
    c|}
    \hline
    \multicolumn{1}{|c|}{\normalsize\textbf{Model}} &
    \multicolumn{1}{c|}{\normalsize\textbf{Provider}} &
    \multicolumn{1}{c|}{\normalsize\textbf{Max tokens}} \\
    \hline

    GPT-4o & Azure Foundry & 4096 \\
    \hline
    GPT-5.2 & Azure Foundry & 16384 \\
    \hline
    Kimi-K2.5 & Azure Foundry & 16384 \\
    \hline
    Claude Opus 4.5 & OpenRouter & 16384 \\
    \hline
    Gemini 3 Pro Preview & OpenRouter & 16384 \\
    \hline
    Qwen3-VL-235B-A22B (Thinking) & OpenRouter & 16384 \\
    \hline
    GLM-4.6V & OpenRouter & 16384 \\
    \hline
    Mistral-Large-3 & OpenRouter & 4096 \\
    \hline
    Llama-4 Maverick-17B-128E (FP8) & OpenRouter & 4096 \\
    \hline

    \end{tabularx}
    \caption{\textbf{Inference settings used in evaluation.}
    All models were evaluated with temperature set to 0. For models that support a reasoning-effort setting, we always use the \texttt{high} level.}
    \label{tab:model-hyperparams}
\end{table}

\subsection{Full Results}
\label{apdx:full-results}

\autoref{tab:full-results} reports the median success rate for each model under all evaluated experimental conditions. Due to cost constraints, GPT 5.2, Claude Opus 4.5 and Gemini 3 Pro Preview were not evaluated under all ablation conditions; missing entries are marked with ``--''.

\begin{table}[H]
    \centering
    \scriptsize
    \renewcommand{\arraystretch}{1.4}
    \setlength{\tabcolsep}{3pt}
    \begin{tabular}{|
    >{\raggedright\arraybackslash}p{2.3cm}|
    c|c|c|c|c|c|c|c|c|}
    \hline
    \multicolumn{1}{|c|}{\footnotesize\textbf{Model}} &
    \rotatebox{70}{\footnotesize\textbf{Baseline}} &
    \rotatebox{70}{\footnotesize\textbf{Detailed}} &
    \rotatebox{70}{\footnotesize\textbf{No Feedback}} &
    \rotatebox{70}{\footnotesize\textbf{No Hand}} &
    \rotatebox{70}{\footnotesize\textbf{No Memory}} &
    \rotatebox{70}{\footnotesize\textbf{No Prev.\ Image}} &
    \rotatebox{70}{\footnotesize\textbf{Text Only ($F_s$)}} &
    \rotatebox{70}{\footnotesize\textbf{Text Only ($F_d$)}} &
    \rotatebox{70}{\footnotesize\textbf{Text Only ($F_n$)}} \\
    \hline

    Claude Opus 4.5      & 76.5 &  --  &  --   &  --  &  --  & 77.8  & 36.1 & --   & -- \\
    \hline
    Gemini 3 Pro         & 72.5 &  --  &  --   &  --  &  --  & 78.7  & 35.2 & --   & -- \\
    \hline
    GPT-5.2              & 71.3 &  --  &  --   &  --  &  --  & --    & 25.0 & --   & -- \\
    \hline
    Kimi-K2.5            & 68.8 & 66.4 & 55.2  & 61.4 & 61.1 & 64.8  & 21.9 & 41.4 & 14.2 \\
    \hline
    Qwen3-VL-235B        & 32.4 & 35.2 & 26.9  & 24.1 & 29.6 & 31.5  & 13.9 & 38.3 & 8.3 \\
    \hline
    GLM-4.6V             & 30.6 & 36.1 & 23.1  & 15.7 & 27.8 & 23.1  & 8.3  & 13.9 & 2.8 \\
    \hline
    GPT-4o               & 23.4 & 41.2 & 21.3  & 22.7 & 29.9 & 18.3  & 4.2  & 8.6  & 0.0 \\
    \hline
    Mistral-Large-3      & 7.4  & 11.8 & 7.2   & 2.5  & 12.5 & 8.3   & 7.9  & 16.9 & 3.0 \\
    \hline
    Llama-4 Maverick     & 5.8  & 11.1 & 7.2   & 5.3  & 9.0  & 8.3   & 6.7  & 9.5  & 4.6 \\
    \hline

    \end{tabular}
    \caption{\textbf{Median success rates (\%) across all experimental conditions.}
    All values are medians over N independent runs (see \autoref{tab:model-hyperparams} for per-model run counts). ``--'' indicates conditions not evaluated for that model due to budget constraints. The three \textit{Text Only} conditions use text-only observations (no images) with different feedback modes: $F_s$ = simple feedback, $F_d$ = detailed feedback, $F_n$ = no feedback.}
    \label{tab:full-results}
\end{table}

\subsection{Sample Visual Errors}
\label{apdx:sample-visual-errors}

\begin{figure}[H]
    \centering
    \small
    \renewcommand{\arraystretch}{1}
    \begin{tabularx}{\textwidth}{|>{\centering\arraybackslash}p{0.2\textwidth}|>{\arraybackslash}X|}
    \hline
    \multicolumn{1}{|c|}{\normalsize\textbf{Observation}} & \multicolumn{1}{c|}{\normalsize\textbf{Model Output}} \\
    \hline

    \adjustimage{width=\linewidth, margin={0pt 6pt 0pt 6pt}, valign=t}{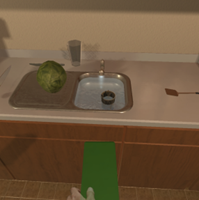} &
    \vspace{3em}\textit{I am holding something green at the bottom center of the view; it looks like a Stool seat/back. Based on the visible object in-hand, I am holding the Stool.} \\
    \hline

    \adjustimage{width=\linewidth, margin={0pt 6pt 0pt 6pt}, valign=t}{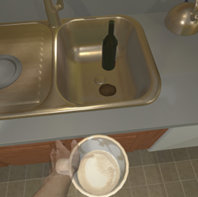} &
    \vspace{3em}\textit{A Mug is visible in my hand at the bottom center; it contains a light-brown liquid (coffee).} \\
    \hline

    \adjustimage{width=\linewidth, margin={0pt 6pt 0pt 6pt}, valign=t}{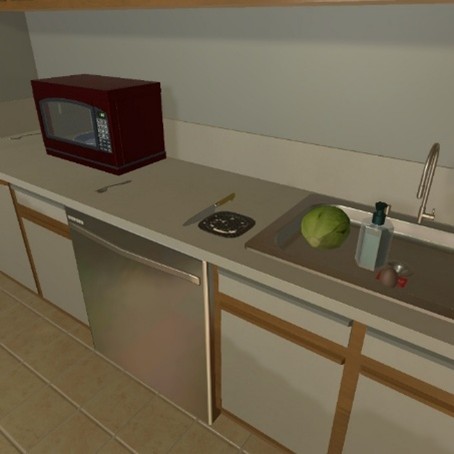} &
    \vspace{4em}\textit{I do not see a Toaster or a Plate in this view.} \\
    \hline

    \adjustimage{width=\linewidth, margin={0pt 6pt 0pt 6pt}, valign=t}{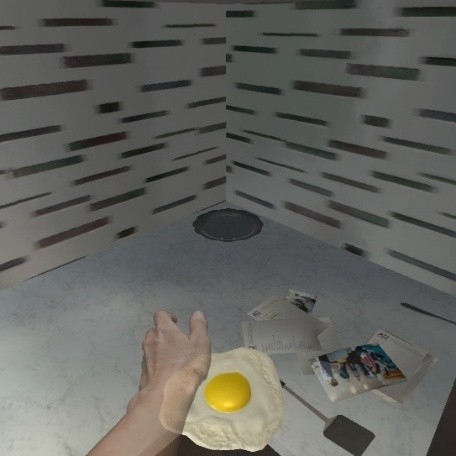} &
    \vspace{4em}\textit{I do not see a Plate, Pan, or StoveBurner in this view.} \\
    \hline

    \end{tabularx}
    \caption{\textbf{Examples of GPT-5.2 visual interpretation errors.}
    Each row shows a visual observation (left) paired with the model's description of what it sees (right). These examples highlight common failure modes where GPT-5.2 misidentifies object states or scene configurations despite clear visual evidence.}
    \label{fig:sample-visual-errors}
\end{figure}

\autoref{fig:sample-visual-errors} shows some additional examples of visual errors (using GPT-5.2).

\subsection{Prompt Example}
\label{apdx:prompt-example}

Note that in the following prompt, the last two in-context examples have been omitted for brevity.

\begin{Verbatim}[breaklines]
You are acting inside a simulated 3D environment.
Your task is: Prepare a mug of coffee, drink it, then wash and store the mug.

------------------------------------------------------------
SECTION: IMAGE GUIDANCE
------------------------------------------------------------
- The first image shows the state of the world before the last action was attempted.
- The second image shows the current state of the world (after the last action was attempted).
- Base all your scene observations on the SECOND image (current state).

------------------------------------------------------------
SECTION: AVAILABLE OBJECTS
------------------------------------------------------------
Apple, AppleSliced, Bowl, Bread, BreadSliced, ButterKnife, Cabinet, Chair, CoffeeMachine, CounterTop, Cup, DishSponge, Egg, EggCracked, Faucet, Floor, Fork, Fridge, GarbageCan, Knife, Ladle, Lettuce, LettuceSliced, LightSwitch, Microwave, Mug, Pan, Plate, Pot, Potato, PotatoSliced, SaltShaker, Sink, SinkBasin, SoapBottle, Spatula, Spoon, StoveBurner, Toaster, Tomato, TomatoSliced, Window

------------------------------------------------------------
SECTION: AVAILABLE ACTIONS
------------------------------------------------------------

Each action consists of exactly two words: ACTION and OBJECT, where OBJECT is one of the AVAILABLE OBJECTS.

- CLEAN {object_name}        : Clean {object_name}
- CLOSE {container_name}     : Close {container_name}
- EMPTY {container_name}     : Pour out liquid from {container_name} (liquids only, not solid items)
- DRINK {container_name}     : Drink liquid from {container_name}
- FIND {object_name}         : Look around to find {object_name}
- OPEN {container_name}      : Open {container_name}
- PICKUP {object_name}       : Pick up {object_name} and hold it
- PUT {container_name}       : Put held item into {container_name}
- SLICE {object_name}        : Slice {object_name} (or crack Egg)
- SPRAY {object_name}        : Spray {object_name} with held object
- TOGGLE_ON {object_name}    : Turn on {object_name}
- TOGGLE_OFF {object_name}   : Turn off {object_name}

-------------------------------------
SECTION: WORLD RULES
--------------------------------------

Follow these rules when reasoning or performing actions:

# OBJECTS #

CoffeeMachine:
- To make coffee, TOGGLE_ON and then TOGGLE_OFF, no need to wait.
- After drinking, clean the Mug before putting it away.

Toaster:
- TOGGLE_ON to start, then TOGGLE_OFF, no need to wait.

Microwave:
- To cook, TOGGLE_ON and then TOGGLE_OFF, no need to wait.
- CLOSE it after removing the cooked item.

StoveBurner:
- Items to cook must be in a Pan
- To cook, TOGGLE_ON and then TOGGLE_OFF, no need to wait.

Fridge:
- Unless being used for your task, when putting food items away, place them inside the fridge

Television:
- To TOGGLE_ON or TOGGLE_OFF Television use the RemoteControl.

Cabinet and Drawer:
- You must CLOSE the Cabinet or Drawer after completing PICKUP or PUT.

Egg:
- Use SLICE to turn an Egg into an EggCracked (it does NOT require a knife)
- The Egg must be PUT on a Pan before you use SLICE
- To cook an EggCracked it must be in a Pan
- A cooked EggCracked must be served on a Plate

Potato:
- Whole Potatoes are cooked in the Microwave
- A cooked Potato must be served in a Bowl

PotatoSliced:
- To cook a PotatoSliced it must be PUT in a Pan
- A PotatoSliced is served on a Plate

Mirror:
- To clean the Mirror, first SPRAY Mirror with SprayBottle, then CLEAN while holding the Cloth.

# ACTIONS #

SLICE:
- You must be holding the Knife to SLICE an object (except for Eggs, which can be SLICED without a Knife).
- You cannot SLICE an object that you are holding
- Valid objects to SLICE: Apple, Bread, Egg, Lettuce, Potato, Tomato.
- SLICE actions produce new objects (e.g., Bread -> BreadSliced) that may move out of view.
- Item must be on a flat surface, CounterTop, DiningTable, etc. (or a Pan for Eggs) to SLICE.
- Items on lower shelves must be moved before you can SLICE them.

PUT:
- The container must be clean
- Specify destination when PUTTING (e.g., PUT SinkBasin).

PICKUP:
- You can only be holding *one* item at a time. You cannot PICKUP a new object while holding another; PUT the held object first.

Serving Food:
- PotatoSliced, EggCracked, BreadSliced, TomatoSliced, AppleSliced are served on a Plate.
- Potato, LettuceSliced are served in a Bowl.

Cooking:
- Pan must be placed on StoveBurner before you can TOGGLE_ON the StoveBurner.
- Using TOGGLE_ON will cook the item in the Pan.
- After it is cooked, TOGGLE_OFF the StoveBurner.

Storage:
- Fruit and vegetables -> Fridge
- Silverware (Knife, Fork, Spoon, ButterKnife, Spatula, Ladle) -> Drawer
- Dishes (Bowl, Plate, Mug) -> Cabinet
- Cookware (Pan, Pot, Kettle) -> CounterTop

Washing Items:
- The Faucet can only be turned on if the SinkBasin is empty OR contains only dishes/cookware.
- Other items like food, or utensils will block the Faucet from turning on.
- Before TOGGLE_ON Faucet, ensure SinkBasin contains only the item you want to wash.
1. PUT the object in the SinkBasin.
2. PICKUP the DishSponge.
3. TOGGLE_ON the Faucet.
4. CLEAN the object (must be holding the DishSponge and the object must be in the sink).
5. TOGGLE_OFF the Faucet.


General:
- All items MUST be CLEAN before being PUT away.
- If a container (Bowl, Plate, Pot) or item is dirty, it MUST be washed before you can PUT items inside it

------------------------------------------------------------
SECTION: AGENT NOTES
------------------------------------------------------------

HISTORY OF ATTEMPTED ACTIONS:
0. FIND Mug  Success
1. PICKUP Mug  Success
2. DRINK Mug  Failure
3. DRINK Mug  Failure
4. FIND SinkBasin  Success
5. DRINK Mug  Failure
6. DRINK Mug  Failure
7. DRINK Mug  Failure
8. PUT CounterTop  Success
9. PICKUP Egg  Success
10. PUT CounterTop  Success
11. PUT SinkBasin  Failure
12. PICKUP Mug  Success
13. PUT SinkBasin  Success
14. PICKUP SaltShaker  Success
15. PUT CounterTop  Success
16. FIND DishSponge  Success
17. PICKUP DishSponge  Success
18. FIND SinkBasin  Success
19. TOGGLE_ON Faucet  Success
20. CLEAN Mug  Success
21. TOGGLE_OFF Faucet  Success


SUGGESTED PLAN SEQUENCE (from your previous response):
1. TOGGLE_OFF Faucet
2. FIND SinkBasin
3. PICKUP Mug
4. FIND CoffeeMachine
5. PUT CoffeeMachine
6. TOGGLE_ON CoffeeMachine
7. TOGGLE_OFF CoffeeMachine
8. FIND Mug
9. PICKUP Mug
10. DRINK Mug
11. FIND SinkBasin
12. PUT SinkBasin
13. FIND DishSponge
14. PICKUP DishSponge
15. FIND SinkBasin
16. TOGGLE_ON Faucet
17. CLEAN Mug
18. TOGGLE_OFF Faucet
19. FIND CounterTop
20. PUT CounterTop
21. FIND Cabinet
22. OPEN Cabinet
23. FIND Mug
24. PICKUP Mug
25. FIND Cabinet
26. PUT Cabinet
27. CLOSE Cabinet

NOTE: This was YOUR proposed plan from the previous step. Evaluate whether it still makes sense given what you observe now. Do NOT follow it blindly - adapt based on the current state.

------------------------------------------------------------
SECTION: TASK INSTRUCTIONS
------------------------------------------------------------
1. Provide detailed reasoning between <think> and </think>.
    - Include the following sections
    VISUAL OBSERVATIONS:
        - List what you observe in the current (second) image:
            - Base your observations on visual appearance only, do NOT use SUGGESTED PLAN SEQUENCE.
            - List objects from AVAILABLE OBJECTS that you see
            - Note containment: What objects are inside or on top of other objects
              (e.g., "Apple is in the Sink", "Mug is on the CounterTop", "Plate contains Egg")
            - Note object states if visible: clean/dirty, sliced/whole, cooked/raw, on/off, open/closed
    INVENTORY CHECK:
        - Determine if you are holding something and what the OBJECT is
        - Use your VISUAL OBSERVATIONS and ACTION HISTORY FEEDBACK
        - If you see an object in your hand or fingers, you ARE holding it (state what it is)
        - If you are holding something then your hand or fingers and the object will be visible at the bottom center of the current (second) image
        - If you see NO hand/fingers or object at the bottom center, you are NOT holding anything
        - CAUTION: Objects resting on surfaces near the bottom of the frame can be mistaken for held objects.
          - When the visual is ambiguous use ACTION HISTORY FEEDBACK
    LAST ACTION ASSESSMENT:
        - Treat ACTION HISTORY FEEDBACK as ground truth for action success or failure
            - Ground your assessment by comparing the first and second images
        - If the action FAILED, analyze WHY:
             - Which WORLD RULE was violated or which precondition was not met?
    DOUBLE CHECK:
        - Do not assume that everything from SUGGESTED PLAN SEQUENCE is accurate, things may have changed.
        - Evaluate if SUGGESTED PLAN SEQUENCE is still valid.
            - Check each item against the visual evidence in the current (second) image. If it doesn't align, update your understanding.
            - For example, if an object was assumed to be clean but appears dirty in the current (second) image, adjust your reasoning accordingly.
            - If the SUGGESTED PLAN SEQUENCE mentions an object you don't see, it may be because you are looking somewhere else
                 - Assume steps about unseen objects are valid, unless there was a state change like cooking or slicing.
    PLAN REASONING:
       - Based on your assessments above, and the WORLD RULES, reason about what to do next to progress toward completing the task.
       - Treat this section as a free-form reasoning space to work through your thoughts.
       - Reason as if you were a human thinking deeply about the task.
2. Provide your output between <answer> and </answer>.
   - Include the following sections
    SUGGESTED PLAN SEQUENCE
        - Include a section titled "SUGGESTED PLAN SEQUENCE:" with a numbered list of ALL remaining actions needed to complete the task.
            - Each line should be: N. $ACTION $OBJECT where $ACTION in AVAILABLE ACTIONS and $OBJECT in AVAILABLE OBJECTS.
            - Start from your NEXT action (the one you're about to take) through task completion.
            - If a SUGGESTED PLAN SEQUENCE was provided from your previous response, evaluate it against current observations.
            - Do NOT blindly copy the previous plan - adapt it based on what you actually see now.
            - If actions succeeded or failed unexpectedly, update the plan accordingly.
        - NEVER repeat the same $ACTION $OBJECT that you did in the last turn - choose a new $ACTION or $OBJECT.
    THINGS TO REMEMBER:
        - Include a section titled "THINGS TO REMEMBER:" with a bulleted list (each starting with "- ") of key facts to retain for future steps.
        - IMPORTANT: What you output in your memory section will be read AFTER your action executes, but the action might fail. Write memories defensively:
            - For example: do NOT write "Currently holding X" if your action is PUT - write instead: "Attempting PUT X - if successful, hands will be empty"
        - Do not assume that everything from SUGGESTED PLAN SEQUENCE should be maintained, use your DOUBLE CHECK from above.
        - Items in "THINGS TO REMEMBER:" should come from the VISUAL OBSERVATIONS above or DOUBLE CHECK, not from assumptions.
    ACTION:
        - Before choosing your action, verify all preconditions are met for that ACTION
        - End the answer with: Action: $ACTION $OBJECT where $ACTION in AVAILABLE ACTIONS and $OBJECT in AVAILABLE OBJECTS.

------------------------------------------------------------
SECTION: EXAMPLES
------------------------------------------------------------

Example 1:
Your primary goal is: Drink a mug coffee then clean and put the mug away

HISTORY OF ATTEMPTED ACTIONS:
0. FIND Mug  Success
1. PICKUP Mug  Success
2. DRINK Mug  Failure

SUGGESTED PLAN SEQUENCE (from your previous response):
1. FIND SinkBasin
2. PUT SinkBasin
3. FIND DishSponge
4. PICKUP DishSponge
5. FIND SinkBasin
6. TOGGLE_ON Faucet
7. CLEAN Mug
8. TOGGLE_OFF Faucet
9. FIND CounterTop
10. PUT CounterTop
11. FIND Cabinet
12. OPEN Cabinet
13. FIND Mug
14. PICKUP Mug
15. FIND Cabinet
16. PUT Cabinet
17. CLOSE Cabinet

Response:

<think>
VISUAL OBSERVATIONS:
- I see a CoffeeMachine sitting on the CounterTop.
- I see a Fork on the CounterTop.
- I see part of a Pan on the right edge of the CounterTop.
- No Sink/SinkBasin/Faucet/Cabinet visible in this frame.
- I see a Mug held in the agent's hand at the bottom center of the view (looks empty from this angle; no visible coffee surface).

INVENTORY CHECK:
- I am holding the Mug (visible in hand).

LAST ACTION ASSESSMENT:
- Last action was DRINK Mug, Mug is empty but HISTORY OF ATTEMPTED ACTIONS says it failed.
- Likely reason: the Mug does not currently contain coffee (or any drinkable liquid), so DRINK cannot proceed.

DOUBLE CHECK:
- I assume the Mug was found near the CoffeeMachine - I can see the CoffeeMachine in the current frame, this is consistent.
- I assume the Mug was filled with coffee, this does not match my VISUAL OBSERVATIONS as the Mug appears to be empty
- I assume I am holding an empty Mug - this matches my VISUAL OBSERVATIONS (Mug visible in hand, appears empty).

PLAN REASONING:
- To "drink a mug coffee", I need the Mug to contain coffee. The CoffeeMachine is visible, so the next step is to PUT the Mug into the CoffeeMachine, then run it (TOGGLE_ON then TOGGLE_OFF), then PICKUP and DRINK.
- After drinking, I must wash the Mug in the sink using DishSponge + Faucet, then store it in the Cabinet (need to FIND those when ready).
</think>

<answer>
THINGS TO REMEMBER:
- I am holding an empty Mug (no coffee inside).
- CoffeeMachine is visible on the CounterTop.
- Attempting PUT CoffeeMachine - if successful, Mug will be in CoffeeMachine ready for brewing.
- After brewing: PICKUP Mug, DRINK, then wash and put in Cabinet.

SUGGESTED PLAN SEQUENCE:
1. FIND CoffeeMachine
2. PUT CoffeeMachine
3. TOGGLE_ON CoffeeMachine
4. TOGGLE_OFF CoffeeMachine
5. FIND Mug
6. PICKUP Mug
7. DRINK Mug
8. FIND SinkBasin
9. PUT SinkBasin
10. FIND DishSponge
11. PICKUP DishSponge
12. FIND SinkBasin
13. TOGGLE_ON Faucet
14. CLEAN Mug
15. TOGGLE_OFF Faucet
16. FIND CounterTop
17. PUT CounterTop
18. FIND Cabinet
19. OPEN Cabinet
20. FIND Mug
21. PICKUP Mug
22. FIND Cabinet
23. PUT Cabinet
24. CLOSE Cabinet

Action: PUT CoffeeMachine
</answer>

Example 2:
...

Example 3:
...
\end{Verbatim}

\end{document}